\pdfoutput=1
\documentclass[11pt]{article}
\usepackage[final]{acl} 

\usepackage[most]{tcolorbox}
\usepackage{times}
\usepackage{latexsym}
\usepackage{algorithm}

\usepackage{pifont}
\usepackage{algorithmic}
\usepackage{amsthm}
\usepackage[T1]{fontenc}
\usepackage{tcolorbox}

\usepackage[utf8]{inputenc}
\usepackage{microtype}
\usepackage{inconsolata}
\usepackage{makecell}
\usepackage{multirow}
\usepackage{setspace}
\usepackage{booktabs}
\usepackage{threeparttable}
\usepackage{CJKutf8}
\usepackage{enumitem}
\usepackage{enumerate}
\usepackage{amsfonts,amssymb,amsmath} 
\usepackage{color}
\usepackage{graphicx}
\usepackage{tabularx} 
\usepackage{placeins}
\usepackage{float} 
\usepackage{hyperref}

\title{Evaluating Self-Generated Documents for Enhancing Retrieval-Augmented Generation with Large Language Models}

\author{
 \textbf{Jiatao Li\textsuperscript{1,2}},
 \textbf{Xinyu Hu\textsuperscript{1}},
 \textbf{Xunjian Yin\textsuperscript{1}},
 \textbf{Xiaojun Wan\textsuperscript{1}}
\\
 \textsuperscript{1}Wangxuan Institute of Computer Technology, Peking University
\\
 \textsuperscript{2}Information Management Department, Peking University
 \\
 \texttt{{leejames@stu.pku.edu.cn}},
 \texttt{\{huxinyu,xjyin,wanxiaojun\}@pku.edu.cn}
}

\usepackage{tcolorbox}

\tcbuselibrary{skins, breakable, theorems}

\definecolor{myblue}{RGB}{221, 233, 247}
\newcommand{\classbg}[1]{\setlength{\fboxsep}{1pt}\colorbox{myblue}{\textbf{#1}}}

\lstdefinelanguage{mycase}{
    basicstyle=\scriptsize\ttfamily, 
    moredelim = [s][\color{mygrey}]{\{}{\}},
}
\usepackage{listings}
\tcbuselibrary{listings, most}  
\newtcblisting{showcase}[1][]{
  enhanced,
  arc=0em,
  boxrule=.5pt,
  listing only,
  listing options={
    language=mycase,
    upquote=true,
    basicstyle=\scriptsize \ttfamily \setlength{\baselineskip}{1.1\baselineskip},
    breaklines=true,  
    breakindent=0pt,
    xleftmargin=0pt,
    xrightmargin=0pt,
    aboveskip=-4pt,
    belowskip=-4pt,
    columns=fullflexible,
    escapeinside={|}{|},
  },
  colback=white,
  colframe=gray,
  breakable,  
  colbacktitle=gray!10,        
  coltitle=black,              
  attach boxed title to top center={yshift=-3mm},
  #1
}

\newtcolorbox[list inside=prompt,auto counter,number within=section]{prompt}[1][]{
    colbacktitle=black!60,
    coltitle=white,
    fontupper=\footnotesize,
    boxsep=5pt,
    left=0pt,
    right=0pt,
    top=0pt,
    bottom=0pt,
    boxrule=1pt,
    #1,
}

\begin{document}
\maketitle
\begin{abstract}
The integration of documents generated by LLMs themselves (Self-Docs) alongside retrieved documents has emerged as a promising strategy for retrieval-augmented generation systems. However, previous research primarily focuses on optimizing the use of Self-Docs, with their inherent properties remaining underexplored. To bridge this gap, we first investigate the overall effectiveness of Self-Docs, identifying key factors that shape their contribution to RAG performance (RQ1). Building on these insights, we develop a taxonomy grounded in Systemic Functional Linguistics to compare the influence of various Self-Docs categories (RQ2) and explore strategies for combining them with external sources (RQ3). Our findings reveal which types of Self-Docs are most beneficial and offer practical guidelines for leveraging them to achieve significant improvements in knowledge-intensive question answering tasks.
\end{abstract}

\section{Introduction}
Recent advancements in large language models (LLMs) have revolutionized natural language processing (NLP) with their impressive ability to understand and generate human-like text \citep{chang2023surveyevaluationlargelanguage,ram-etal-2023-context,brown2020languagemodelsfewshotlearners,ouyang2022traininglanguagemodelsfollow}. Despite these strides, LLMs remain susceptible to factual inaccuracies and hallucinations, hindering their real-world reliability \citep{maynez-etal-2020-faithfulness,zhou-etal-2021-detecting,Ji_2023}. Retrieval-augmented generation (RAG) has emerged as a promising solution by incorporating external knowledge into the generation process, thereby grounding responses in verifiable information \citep{NEURIPS2020_6b493230,ram-etal-2023-context,kandpal2023largelanguagemodelsstruggle}.

While retrieval systems significantly enhance LLM capabilities, recent research points to the potential of Self-Docs as a powerful addition to the RAG framework \citep{zhang2023merginggeneratedretrievedknowledge,liu-etal-2022-generated,sun2023recitationaugmentedlanguagemodels,yu2023generateretrievelargelanguage}. Self-Docs, autonomously generated by the model, offer a dynamic way to supplement retrieved content and enrich the information available for downstream tasks. 

Despite the growing interest in Self-Docs, previous related work has mostly focused on how to better utilize their information~\cite{liu2022generatedknowledgepromptingcommonsense,fang2022leveraging,yu2023generateretrievelargelanguage}, such as designing specific methods to handle conflicting knowledge~\cite{zhang2023merginggeneratedretrievedknowledge}, while neglecting the impact of the inherent features of Self-Docs. Intuitively, if we have a more thorough understanding of these features and know which ones are more likely to improve the performance of RAG, the potential of Self-Docs will be fundamentally increased. This paper aims to present a comprehensive exploration of Self-Docs within RAG by exploring three progressively posed research questions:

\begin{itemize}[leftmargin=*,itemsep=1pt]
\item RQ1: To what extent do Self-Docs enhance the performance of RAG? 
\item RQ2: Which types of Self-Docs are most effective across different scenarios? 
\item RQ3: How can Self-Docs be effectively integrated with external sources (e.g., Wikipedia) based on their properties to further enhance the performance? 
\end{itemize}

\begin{figure*}[h]
  \centering
  \includegraphics[width=\textwidth]{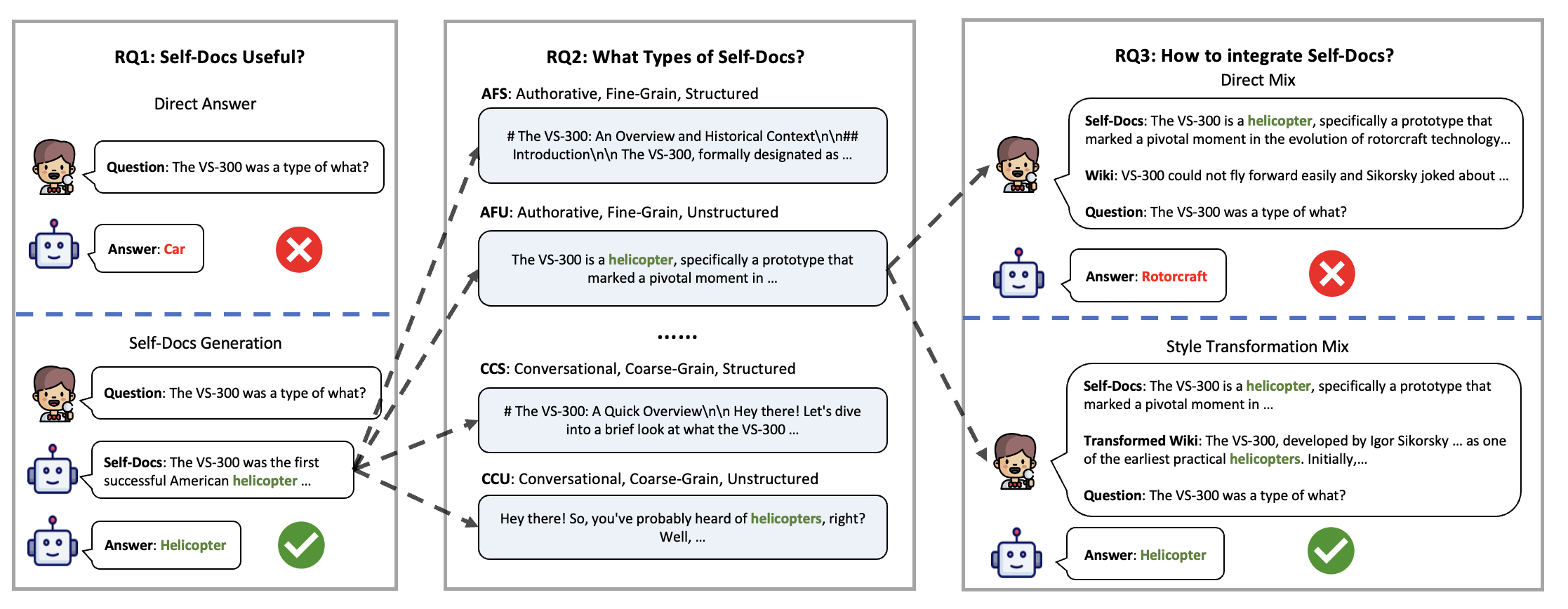}
  \caption{An illustrative overview of our three-stage research framework. RQ1 assesses the baseline utility of Self-Docs in RAG, RQ2 identifies which Self-Doc types are most effective for different tasks, and RQ3 demonstrates how integrating these optimal Self-Docs with external knowledge sources—especially through stylistic alignment—further enhances performance.}
  \label{fig:overview}
\end{figure*} 

As illustrated in Figure~\ref{fig:overview}, our investigation begins by assessing the overall impact of Self-Docs on RAG performance (RQ1). In this initial phase, we consider how factors such as task type, the number of documents, the size of the base large language model (LLM), and the source of the documents influence performance gains. Building on the insights gained here, we then focus on identifying which types of Self-Docs excel in particular scenarios (RQ2). To guide this analysis, we draw upon Systemic Functional Linguistics (SFL) theory, which offers a structured framework for understanding language through its ideational, interpersonal, and textual metafunctions. By applying SFL principles, we categorize Self-Docs according to their granularity, tone, and structure, linking these qualities to distinct communicative functions and enabling us to pinpoint the document characteristics that lead to optimal outcomes.

Finally, armed with a clear understanding of which Self-Docs are most effective, we turn our attention to integrating them with external sources, such as Wikipedia, to further enhance RAG performance (RQ3). We consider both direct integration—simply combining Self-Docs and Wikipedia documents—and a style-transformation approach, which involves aligning the stylistic features of Wikipedia entries with the most effective types of Self-Docs before mixing them.

Our results show that scaling model capacity improves performance, although marginal returns may diminish for complex tasks. Tailoring the quantity, tone, structure, and granularity of Self-Docs to match the task proves essential for maximizing gains. Moreover, harmoniously integrating Self-Docs with external knowledge sources—especially when stylistically aligned—yields more robust improvements than relying on either approach alone. In doing so, this work highlights strategic design choices for Self-Docs and integration methods that lead to superior RAG outcomes across diverse, knowledge-intensive NLP scenarios.

To the best of our knowledge, this is the first work to deeply analyze the characteristics of self-generated documents (Self-Docs) and their effects on retrieval-augmented generation (RAG). Our contributions are as follows:

\begin{itemize}[leftmargin=*,itemsep=1pt]
\item We validate the effectiveness of Self-Docs and uncover key factors influencing their utility in RAG.
\item We assess the effectiveness of different Self-Docs types across open-domain QA, multi-hop QA, long-form QA, and fact verification tasks.
\item We establish a classification framework grounded in Systemic Functional Linguistics (SFL) theory, categorizing documents by their roles, linguistic characteristics, and impact on RAG.
\item We provide actionable insights for incorporating Self-Docs into RAG workflows, demonstrating how selecting the right document types can enhance both accuracy and efficiency.\footnote{Our code and data are released at \url{https://github.com/leejamesss/Eval_Self_Docs_RAG} 
to facilitate related research.}
\end{itemize}

\section{Background}
\subsection{Task Formulation}  

We formally describe the task formulation as follows. The Question Answering (QA) task involves generating a response \( r \) to a given query \( q \), utilizing knowledge \( K \) when available. The knowledge \( K \) can be derived from two primary sources: (1) \textbf{External Knowledge}, such as structured or unstructured repositories (e.g., Wikipedia), or (2) \textbf{Internal Knowledge}, stored within the parametric memory of a large language model (LLM).  

The response generation process is defined as:  
\[
r = \mathcal{L}_{\text{QA}}(q, F(K)),
\]  
where \(\mathcal{L}_{\text{QA}}\) denotes the LLM used for answering the query, and \( F(K) \) represents a post-processing function applied to the knowledge \( K \). Post-processing strategies may include methods like knowledge mixing or style transformation, ensuring the retrieved or generated information aligns with the task requirements.  

\subsection{Definition of Self-Docs}  
Self-Generated Documents (Self-Docs) refer to documents generated solely using the parametric memory of the same LLM for both document generation and downstream QA tasks. Formally, given a query \( q \) and an optional set of instructions \( I \) (e.g., few-shot examples or style-specific prompts), an Self-Docs \( g \) is generated as:  
\[
g = \mathcal{L}_{\text{Self-Docs}}(q, I, \varnothing),
\]  
where \(\mathcal{L}_{\text{Self-Docs}} = \mathcal{L}_{\text{QA}}\) denotes the same LLM used for both document generation and question answering, \( I \) represents task-specific instructions or few-shot examples, and \(\varnothing\) indicates that no external knowledge \( K \) is utilized in the generation process. This formulation ensures that \( g \) is derived entirely from the internal knowledge of the LLM, maintaining consistency between document generation and QA.  

For downstream QA tasks, the response \( r \) is then produced using the same LLM and the generated Self-Docs as knowledge, defined as:  
\[
r = \mathcal{L}_{\text{QA}}(q, F(G)),
\]  
where \( G = \{g_1, g_2, \ldots, g_n\} \) is the set of Self-Docs, and \( F(G) \) represents any post-processing or integration strategy applied to the Self-Docs. By ensuring that \(\mathcal{L}_{\text{Self-Docs}} = \mathcal{L}_{\text{QA}}\), the approach guarantees seamless integration between document generation and QA, leveraging the LLM’s parametric memory while supporting task-specific adaptability.

\subsection{Self-Docs Generation Pipeline}
To generate Self-Docs, we employ the \textbf{GenRead}~\cite{yu2023generateretrievelargelanguage} few-shot prompting method. The process begins with input and clustering, where given a set of queries \( Q = \{q_1, q_2, \ldots, q_n\} \) and their corresponding top-1 retrieved wiki documents \( D = \{d_1, d_2, \ldots, d_n\} \), embeddings are computed for each \((q_i + d_i)\) pair. These embeddings are then clustered into \( k \) clusters \( C = \{C_1, \ldots, C_k\} \) using \( k \)-means. From each cluster \( C_i \), \( m \) query-document pairs are sampled to create few-shot examples in the format \(\text{FewShot}_j = \langle \text{Input: } q_j, \text{Output: } d_j \rangle\). Using these examples, \( k \) Self-Docs are generated for each query \( q_t \) as \( g_{t,i} = \mathcal{L}(q_t \mid \text{FewShots}_i) \), where \(\mathcal{L}\) represents the large language model. The final output consists of all generated documents \( G = \bigcup_{t=1}^n G_t = \{g_{1,1}, \ldots, g_{n,k}\} \).

The detailed pseudocode for the Self-Docs generation process is provided in Appendix~\ref{alg:sgd_pipeline}, and the implementation specifics are described in Appendix~\ref{sec:implement_details}.

\subsection{Systemic Functional Linguistics Theory}
\label{sec:sfl}

To build an effective classification system for Self-Docs that enhances RAG systems, we ground our taxonomy in Systemic Functional Linguistics (SFL). SFL, developed by Michael Halliday~\cite{Halliday1989LanguageCA}, views language as a dynamic tool shaped by social contexts, emphasizing its role in meaning-making rather than static grammatical rules. This framework helps us categorize Self-Docs based on their communicative functions, aligning document types with specific retrieval and generation tasks to improve RAG performance.

At the heart of SFL are three metafunctions —ideational, interpersonal, and textual— which operate together to convey meaning~\cite{inbook,article_bilal}. These metafunctions guide how language expresses ideas, manages social relations, and structures discourse.

\textbf{Ideational Metafunction} focuses on how language represents knowledge and organizes experiences~\cite{article1,article_huta,article_kous}. This includes everything from detailed factual information to abstract concepts, depending on the communicative goal.
  
\textbf{Interpersonal Metafunction} handles the relational dynamics of communication, including tone, authority, and interaction with the audience~\cite{phdthesis,Cheng+2023,article_Chueasuai}. It defines whether the document is formal or conversational and shapes how trust or authority is established.

\textbf{Textual Metafunction} ensures the logical flow and coherence of the document, structuring information for clarity and accessibility~\cite{article_sameer,article_Matthiessen,Koutchadé_2017}. This function is crucial for guiding readers through complex or information-dense content.

By applying SFL to Self-Docs, we can classify documents not only based on the information they provide but also on how they engage readers and structure their content. This approach helps develop a nuanced taxonomy that aligns with the varying communicative demands of different tasks in RAG systems.

\section{Datasets and LLMs}
We conduct zero-shot evaluations following prior work \citep{yu2024rankragunifyingcontextranking}, limiting each dataset to the first 500 samples to reduce computational overhead\citep{trivedi2022interleaving}. Our experiments span four well-established datasets representing different knowledge-intensive tasks: TriviaQA (open-domain QA), HotpotQA (multi-hop QA), FEVER (fact verification), and ELI5 (long-form QA). Table~\ref{tab:setting} summarizes these datasets, their task types, and evaluation metrics. Additional details can be found in Appendix~\ref{sec:dataset_details}.

\begin{figure*}[h]
  \centering
  \includegraphics[width=\textwidth]{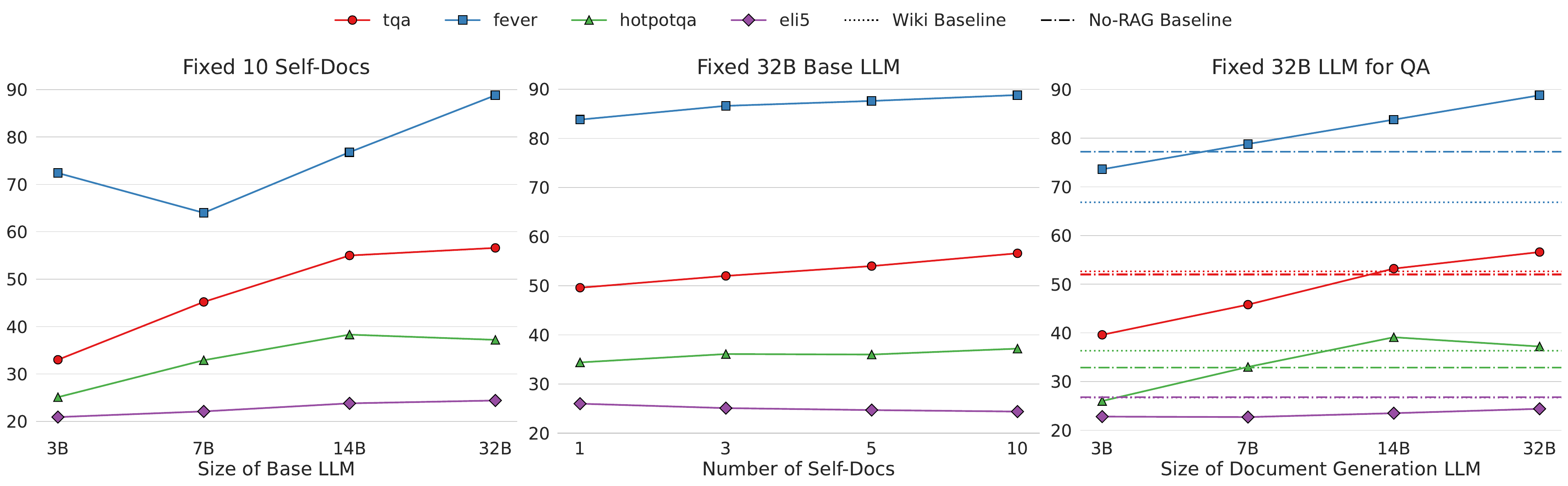}
  \caption{Performance factor analysis examining the effects of base model size, document count, and document generation model size on RAG performance across different tasks. \textbf{Left:} Impact of varying the base model size when using the same Qwen2.5 model for both document generation and question answering, with the number of self-generated documents fixed at 10. \textbf{Middle:} Impact of varying the number of self-generated documents, using the Qwen2.5-32B-Instruct model as the base model. \textbf{Right:} Impact of varying the document generation model size while keeping the question answering model fixed at Qwen2.5-32B-Instruct and the document count at (n=10).}
  \label{fig:rq1}
\end{figure*}

To investigate the impact of model size on the effectiveness of Self-Docs (Section~\ref{sec:rq1}), we use the Qwen2.5 series \citep{qwen2.5}. This family spans a broad range of parameter sizes (0.5B to 72B), enabling a more comprehensive analysis compared to models like LLaMA, which offer fewer parameter variants \citep{dubey2024llama3herdmodels}. Moreover, Qwen2.5 models consistently outperform LLaMA 3.1 models at comparable scales \citep{qwen2.5}.

For controlled experiments, we use only Qwen2.5 models. We focus on the 3B, 7B, 14B, and 32B variants, balancing computational feasibility with capability. Unless otherwise noted, we adopt the largest feasible model—Qwen2.5-32B-Instruct—as the backbone for QA and document generation. To isolate the effect of the document generation model, we hold the QA model fixed at Qwen2.5-32B-Instruct while varying the size of the document generation model. Figure~\ref{fig:rq1} illustrates these configurations.

\begin{table}[t]
\small
\centering
\begin{tabular}{lcc}
\toprule
\textbf{Dataset} & \textbf{Task} & \textbf{Metric} \\
\midrule
\textbf{TQA} & Open-Domain QA & EM \\
\textbf{HotpotQA} & Multi-Hop QA & F$_1$ \\
\textbf{FEVER} & Fact Verification & EM \\
\textbf{ELI5} & Long-Form QA & F$_1$ \\
\bottomrule
\end{tabular}
\caption{Summary of datasets, task types, and evaluation metrics for TQA, HotpotQA, FEVER, and ELI5.}
\label{tab:setting}
\end{table}
\section{Impact of Self-Generated Documents}
\label{sec:rq1}

To understand how Self-Docs influence RAG performance, we consider three key factors: (1) the size of the base model, (2) the number of Self-Docs, and (3) the size of the document generation model.\footnote{Here, the \textit{base model} denotes the LLM used for both question answering and generating Self-Docs, while the \textit{document generation model} refers to an LLM used solely for creating documents.} We employ Qwen2.5 Instruct models of varying sizes (3B, 7B, 14B, and 32B) and test them on four knowledge-intensive tasks (Table~\ref{tab:setting}). Detailed results are provided in Appendix~\ref{sec:detailed_results}.

\subsection{Scaling the Base Model}
Larger base models generally improve results, but these gains diminish at very high scales, particularly for tasks demanding complex reasoning (e.g., HotpotQA). This suggests that simply increasing model size does not guarantee uniform improvements.

\subsection{Varying Document Number}
Raising the number of Self-Docs typically boosts performance, yet the optimal quantity depends on the task. Fact-centric tasks (e.g., FEVER) benefit from more documents, whereas creative reasoning tasks (e.g., ELI5) may actually see performance plateau or even dip when overloaded with documents.

\subsection{Scaling the Document Generation Model}
A larger document generation model consistently yields more task-relevant, higher-quality documents, thereby enhancing RAG performance. While previous studies have highlighted the advantages of large LLMs for document creation \citep{yu2023generateretrievelargelanguage}, our findings refine this understanding: these benefits only emerge once the model surpasses a certain size threshold. Smaller models struggle to exceed both the No-RAG baseline and Wikipedia-based retrieval, underscoring that substantial model capacity is crucial for unlocking the full potential of self-generated documents.

\subsection{Key Takeaways}
While Self-Docs enhance RAG in some settings, their effectiveness is not uniform. In certain tasks, Self-Docs match or surpass external sources like Wikipedia, yet in others, the benefits are less pronounced. This nuanced performance profile suggests that not all Self-Docs are created equal. To fully realize their potential, we need to understand \textit{which kinds} of Self-Docs work best for different scenarios.

This observation motivates our next step, RQ2, where we delve deeper into the types of Self-Docs. By analyzing their linguistic and functional characteristics, we aim to identify which categories consistently drive better results and how to strategically design Self-Docs for optimal impact.
\section{Exploration of Self-Docs Types}
\label{sec:rq2}
\begin{table*}[h]
\centering
\small
\begin{tabular}{c|c|c|c|c|c|c|c|c}
\toprule
\textbf{Docs Type} & \textbf{Interpersonal} & \textbf{Ideational} & \textbf{Textual} & \textbf{\textsc{TQA}} & \textbf{\textsc{HotpotQA}} & \textbf{\textsc{FEVER}} & \textbf{\textsc{ELI5}} & \textbf{Average} \\
\midrule
\textsc{AFS} & Authoritative & Fine-Grain & Structured  & 55.4 & 37.5 & 83.4 & 20.9 & 49.3 \\
\textsc{AFU} & Authoritative & Fine-Grain & Unstructured & \underline{\textbf{58.4}} & \underline{\textbf{39.9}} & \underline{\textbf{85.2}} & 22.8 & \underline{\textbf{51.6}} \\
\textsc{ACS} & Authoritative & Coarse-Grain & Structured & 56.8 & 38.7 & 83.6 & 22.0 & 50.3 \\
\textsc{ACU} & Authoritative & Coarse-Grain & Unstructured & 56.4 & 39.6 & 85.0 & 23.5 & 51.1 \\
\textsc{CFS} & Conversational & Fine-Grain & Structured & 57.4 & 36.4 & 81.8 & 24.0 & 49.9 \\
\textsc{CFU} & Conversational & Fine-Grain & Unstructured & 57.8 & 39.5 & 84.2 & 25.1 & \underline{\textbf{51.6}} \\
\textsc{CCS} & Conversational & Coarse-Grain & Structured & 56.0 & 35.2 & 82.0 & 25.4 & 49.6 \\
\textsc{CCU} & Conversational & Coarse-Grain & Unstructured & 56.6 & 38.7 & 83.8 & \underline{26.4} & 51.4 \\
\midrule
\textsc{Wiki} & Authoritative & Flexible & Unstructured & 52.6 & 36.3 & 66.8 & \textbf{26.7} & 45.6 \\
\bottomrule
\end{tabular}
\caption{Performance comparison of various document types across tasks. Each document type is categorized based on its interpersonal, ideational, and textual functions. 
The best overall performance for each dataset is highlighted in \textbf{bold}, and the best performance within the category of Self-Docs is \underline{underlined}.}
\label{tab:rq2_full}
\end{table*}

\begin{table*}[h]
\vspace{-1mm}
\small
\centering
    \begin{tabular}{c|c|cccc|c}
        \toprule
        \textbf{Meta Function} & \textbf{Dimension} & \textbf{\textsc{TQA}} & \textbf{\textsc{HotpotQA}} & \textbf{\textsc{FEVER}} & \textbf{\textsc{ELI5}} & \textbf{Average} \\
        \midrule
        \multirow{2}{*}[0em]{\textsc{Ideational}} 
                           & \textsc{Fine-Grain}      & \underline{57.3} & 
                           \underline{38.3} & \underline{84.3} & 23.2 & \underline{50.8} \\
                           & \textsc{Coarse-Grain}    & 56.4 & 38.1 & 84.0 & \underline{24.3} & 50.7 \\
        \midrule
        \multirow{2}{*}[0em]{\textsc{Interpersonal}} 
                           & \textsc{Authoritative}   & 56.8 & \underline{38.9} & \underline{84.8} & 22.3 & 50.7 \\
                           & \textsc{Conversational}  & \underline{57.0} & 37.5 & 83.6 & \underline{25.2} & \underline{50.8} \\
        \midrule
        \multirow{2}{*}[0em]{\textsc{Textual}} 
                           & \textsc{Structured}      & 56.4 & 37.0 & 83.1 & 23.1 & 49.9  \\
                           & \textsc{Unstructured}    & \underline{57.3} & \underline{39.4} & \underline{85.2} & \underline{24.5} & \underline{51.6} \\
        \bottomrule
    \end{tabular}
\caption{Aggregated performance across different metafunction dimensions (Ideational, Interpersonal, and Textual) for various datasets. Each metafunction dimension is further divided into fine-grain and coarse-grain (for Ideational), authoritative and conversational (for Interpersonal), and structured and unstructured (for Textual). The best performance within each metafunction dimension is \underline{underlined}.}
\vspace{-1mm}
\label{tab:rq2_agg}
\end{table*}
Using the taxonomy from Section~\ref{sec:taxonomy}, we evaluate each Self-Docs type under diverse conditions. Table~\ref{tab:rq2_full} presents their performance per scenario, while Table~\ref{tab:rq2_agg} aggregates results by linguistic and communicative properties. This analysis clarifies which Self-Docs types best support specific tasks, offering strategic insights into more effective document generation.

\subsection{Taxonomy of Self-Generated Documents}
\label{sec:taxonomy}
Self-generated documents (Self-Docs) are classified using Systemic Functional Linguistics (SFL), which identifies three core metafunctions: Ideational, Interpersonal, and Textual, as discussed in Section~\ref{sec:sfl}. The Ideational metafunction focuses on content granularity, distinguishing fine-grained documents like technical papers that provide detailed information from coarse-grained documents like summaries that offer broader overviews. The Interpersonal metafunction examines tone, categorizing documents as authoritative with a formal style (e.g., research papers) or conversational with an engaging tone (e.g., blogs). The Textual metafunction addresses structure, with structured documents like reports providing clarity and easy navigation, while unstructured documents like essays emphasize narrative flow.

These dimensions combine to create eight types of Self-Docs, as shown in Table~\ref{tab:document_types}. Examples include research papers, which are fine-grained, structured, and authoritative, and social media posts, which are coarse-grained, unstructured, and conversational. This classification framework aligns document features with communication goals, enhancing their effectiveness in retrieval-augmented generation systems.

\begin{table}
\centering
\small
\setlength{\tabcolsep}{4pt}
\renewcommand{\arraystretch}{1.25}
\resizebox{\columnwidth}{!}{
\begin{tabular}{ccccc}
\toprule
\textbf{Type} & \textbf{Interpersonal} & \textbf{Ideational} & \textbf{Textual} & \textbf{Example}\\
\midrule
AFS & Authorative & Fine-Grain & Structured  & Research Paper \\
AFU & Authorative & Fine-Grain & Unstructured & Case Studies \\
ACS & Authorative & Coarse-Grain &  Structured & News Article \\
ACU & Authorative & Coarse-Grain & Unstructured & Opinion Pieces \\
CFS & Conversational & Fine-Grain & Structured & How-to Guides \\
CFU & Conversational & Fine-Grain & Unstructured & Personal Essays \\
CCS & Conversational & Coarse-Grain & Structured & Travel Blogs \\
CCU & Conversational & Coarse-Grain & Unstructured & Social Media Posts \\
\bottomrule
\end{tabular}}
\caption{Classification of document types based on the three SFL theory metafunctions: interpersonal, ideational, and textual, including representative examples for each category.}
\label{tab:document_types}
\end{table}

\subsection{Metafunction Analysis}

\paragraph{Textual Metafunction}
Unstructured Self-Docs consistently outperform their structured counterparts across all tasks, achieving the highest average score (Table~\ref{tab:rq2_agg}). It suggests that the flexibility of unstructured content allows the model to better extract and utilize relevant information, potentially enabling more nuanced and adaptable understanding.

\paragraph{Interpersonal Metafunction}
While both authoritative and conversational tones achieve similar average performance, we observe task-dependent differences. Conversational documents are slightly more effective for long-form QA tasks like ELI5, potentially fostering a more engaging and accessible interaction for knowledge exploration. On the other hand, authoritative documents perform slightly better for factual tasks like FEVER, likely due to the emphasis on precision and reliability associated with a formal tone in factual verification scenarios.

\paragraph{Ideational Metafunction}
Fine-grain and coarse-grain documents demonstrate comparable average performance overall. However, we see task-specific variations. Fine-grained documents excel in tasks requiring detailed factual information (e.g., TQA), likely benefiting from the richness of detail and specific references provided. In contrast, the latter perform better in open-ended tasks where broader overviews are more beneficial (e.g., ELI5), potentially due to their ability to provide a more concise and digestible overview of the topic.

\subsection{Task-Specific Analysis}

\noindent\textbf{Fact Verification}\quad
In Fact Verification (FEVER), unstructured, authoritative, and fine-grained documents perform best. This suggests that for tasks requiring high accuracy and reliable information, a combination of content flexibility, a formal tone, and detailed factual content proves most effective.

\noindent\textbf{Multi-hop QA}\quad
For Multi-hop QA (HotpotQA), unstructured documents outperform structured ones, and fine-grained content slightly outperforms coarse-grained content. This indicates that for tasks involving complex reasoning over multiple information sources, a flexible content structure and a higher level of detail are beneficial.

\noindent\textbf{Open-Domain QA}\quad
In Open-Domain QA (TQA), fine-grained and authoritative documents demonstrate a slight advantage. This finding aligns with the need for precision and detailed information in answering factual questions based on a broad knowledge base.

\noindent\textbf{Long-Form QA}\quad
In Long-Form QA (ELI5), conversational and unstructured documents perform best. This suggests that for tasks focused on generating accessible and user-friendly explanations, a more engaging and flexible style is crucial.

\subsection{Key Insights}

\paragraph{Adaptable Formats Improve Performance}
Unstructured documents generally outperform structured ones, suggesting that more adaptable formats help the model better extract and use relevant information.

\paragraph{Tone and Detail Level Are Task-Dependent}
The optimal tone and level of detail depend on the task. For instance, long-form QA (ELI5) benefits from conversational, coarse-grained documents, while fact verification (FEVER) performs best with authoritative, fine-grained content.

\paragraph{Task-Tailored Adjustments Are Crucial}
Tailoring document attributes—such as structure, tone, and granularity—to the specific demands of each task can significantly boost RAG performance. \\

Crucially, these insights set the stage for RQ3, where we investigate how to further enhance performance by integrating Self-Docs with external sources.
\section{Integrating Self-Docs with External Knowledge}
\label{sec:rq3}
\begin{table*}[h]
\centering
\resizebox{\textwidth}{!}{
\begin{tabular}{c|c|c|c|c|c|c|c|c|c}
\toprule
\textbf{Docs Type} & \textbf{Mix Strategy} & \textbf{Interpersonal} & \textbf{Ideational} & \textbf{Textual} & \textbf{\textsc{TQA}} & \textbf{\textsc{HotpotQA}} & \textbf{\textsc{FEVER}} & \textbf{\textsc{ELI5}} & \textbf{Average} \\
\midrule
\textsc{Wiki} & - & Authoritative & Flexible & Unstructured & 52.6  & 36.3  & 66.8  & \textbf{26.7}  & 45.6  \\
\textsc{GenRead mix} & Direct Mix & Authoritative & Flexible & Unstructured & 57.6  & 39.1  & 87.0  & 24.7  & 52.1  \\
\midrule
\textsc{AFU} & Direct Mix & Authoritative & Fine-Grain & Unstructured & 56.8\scriptsize{↓1.6} & 41.0\scriptsize{↑1.1} & 84.4\scriptsize{↓0.8} & 23.1\scriptsize{↑0.3} & 51.3\scriptsize{↓0.3} \\
\textsc{CFU} & Direct Mix & Conversational & Fine-Grain & Unstructured & 57.2\scriptsize{↓0.6} & 41.6\scriptsize{↑2.1} & 85.0\scriptsize{↑0.8} & 25.6\scriptsize{↑0.5} & 52.4\scriptsize{↑0.8} \\
\textsc{CCU} & Direct Mix & Conversational & Coarse-Grain & Unstructured & \underline{57.4}\scriptsize{↑0.8} & \underline{41.8}\scriptsize{↑3.1} & \underline{85.8}\scriptsize{↑2.0} & \underline{26.5}\scriptsize{↑0.1} & \underline{52.9}\scriptsize{↑1.5} \\
\midrule
\textsc{AFU} & Style Transformation Mix & Authoritative & Fine-Grain & Unstructured & \underline{\textbf{59.4}}\scriptsize{↑1.0} & \underline{\textbf{43.3}}\scriptsize{↑3.4} & \underline{\textbf{87.6}}\scriptsize{↑2.4} & 22.7\scriptsize{↓0.1} & 53.3\scriptsize{↑1.7} \\
\textsc{CFU} & Style Transformation Mix & Conversational & Fine-Grain & Unstructured & 58.0\scriptsize{↑0.2} & 41.0\scriptsize{↑1.5} & 86.4\scriptsize{↑2.2} & 25.3\scriptsize{↑0.2} & 52.7\scriptsize{↑1.1} \\
\textsc{CCU} & Style Transformation Mix & Conversational & Coarse-Grain & Unstructured & 57.6\scriptsize{↑1.0} & 43.0\scriptsize{↑4.3} & 87.4\scriptsize{↑3.6} & \underline{26.3}\scriptsize{↓0.1} & \underline{\textbf{53.6}}\scriptsize{↑2.2} \\
\bottomrule
\end{tabular}
}
\caption{Performance comparison across different generated document types combined with Wiki documents using mix with style transformation strategy for various tasks. Superscript values indicate the difference from Table~\ref{tab:rq2_full}, with increases marked by an upward arrow (↑) and decreases by a downward arrow (↓). The best overall performance for each dataset is highlighted in \textbf{bold}, and the best performance within the category of Self-Docs with the same mix strategy is \underline{underlined}.}
\label{tab:rq3_full}
\end{table*}

In this section, we build on the lessons from RQ1 and RQ2. RQ1 showed that scaling models and adjusting document counts can improve performance, though inconsistently, while RQ2 demonstrated that tailoring document attributes to each task’s demands fosters more reliable gains. Now we seek to combine these refined Self-Doc types with external data—such as Wikipedia—ensuring that the additional knowledge is both complementary and stylistically aligned. This approach aims to unlock the full potential of Self-Docs, leading to even more robust and contextually rich RAG performance.


\subsection{Integration Strategies}
We examine two integration strategies: \textit{Direct Mix} and \textit{Style-Transformation Mix}. Table~\ref{tab:rq3_full} presents their performance for top-performing Self-Doc types (as identified in Table~\ref{tab:rq2_full}) across various tasks.
\paragraph{Direct Mix} 
The Direct Mix strategy directly concatenates the top \( k \) ranked Wiki documents, denoted as \( W = \{W_1, W_2, \dots, W_k\} \), with the first \( n \) Self-Docs from a predefined taxonomy, represented as \( S = \{S_1, S_2, \dots, S_n\} \). The final mixed document set \( D_{\text{Mixed}} \) is given by:
$$
\textstyle
D_{\text{Mixed}} = \text{Concat}\left( \bigcup_{i=1}^{k} W_i, \bigcup_{j=1}^{n} S_j \right)
$$
As shown in Table~\ref{tab:rq3_full}, the Direct Mix strategy successfully combines broad factual knowledge from Wiki documents with task-specific insights from Self-Docs. However, performance can be affected by stylistic inconsistencies between the Wiki content and the Self-Docs, which may lead to variability in tone and structure.

\paragraph{Style-Transformation Mix} 
The Style-Transformation Mix strategy addresses the stylistic mismatch by transforming the retrieved Wiki documents to align with the style of the Self-Docs. The transformation function \( T(W_i) \) uses a large language model to adapt the style of each Wiki document \( W_i \) to match the Self-Docs. The final mixed document set is constructed as:
$$
\textstyle
D_{\text{Mixed}} = \text{Concat}\left( \bigcup_{i=1}^{k} T(W_i), \bigcup_{j=1}^{n} S_j \right)
$$

To implement this approach, we extend GenRead~\cite{yu2023generateretrievelargelanguage} with style alignment. This involves transforming the retrieved documents \( d_j \) within the few-shot examples into target-style documents \( t_j \) using a style transformation function, expressed as \( t_j = \text{StyleTransform}(d_j, T) \). The style-aligned examples, formatted as \( \text{FewShot}_j = \langle \text{Input: } q_j, \text{Output: } t_j \rangle \), are then used to generate Self-Docs that conform to the specified style. The final outputs remain as \( G = \bigcup_{t=1}^n G_t = \{g_{1,1}, \ldots, g_{n,k}\} \), ensuring that the generated documents are both content-rich and style-consistent.

\subsection{Performance Comparison}
The results in Table~\ref{tab:rq3_full} demonstrate that integrating Wiki documents with Self-Docs improves performance across different tasks and document types. The Style-Transformation Mix strategy shows clear advantages over Direct Mix, emphasizing the importance of stylistic alignment between external knowledge and Self-Docs for better RAG system performance.

\subsection{Key Insights}

\paragraph{External Integration Improves Performance}
Both Direct Mix and Style-Transformation Mix generally outperform the use of Self-Docs alone, confirming the value of incorporating external knowledge sources.

\paragraph{Stylistic Alignment Drives Gains}
Style-Transformation Mix, in particular, delivers notable improvements by ensuring stylistic consistency. Aligning tone, structure, and granularity increases the effectiveness of external knowledge integration.

\paragraph{Harmonizing Content is Essential}
These results underscore the importance of not only adding external information but also harmonizing its presentation style with self-generated documents to maximize RAG performance.

\section{Related Work}

\paragraph{Generation-Augmented Approaches}
Recent advances in \textbf{generation-augmented methods} leverage LLMs to generate intermediate contexts that enhance task performance by drawing on their pre-trained parametric knowledge \citep{petroni2019languagemodelsknowledgebases, roberts2020knowledgepackparameterslanguage,chen-etal-2023-beyond}. The generated contexts span diverse knowledge domains, including background information \citep{sun2023recitationaugmentedlanguagemodels, yu2023generateretrievelargelanguage}, commonsense reasoning \citep{liu2022generatedknowledgepromptingcommonsense}, and domain-specific expertise \citep{feng2024knowledgecardfillingllms, luo2023augmentedlargelanguagemodels}, along with chain-of-thought reasoning processes \citep{NEURIPS2022_9d560961, kojima2022large}.

\paragraph{Linguistic Frameworks in NLP}
Linguistic frameworks have been employed to improve the coherence of language model generated content \citep{https://doi.org/10.1111/j.1467-8640.1992.tb00339.x, annurev:/content/journals/10.1146/annurev-linguistics-030521-043632, Miaschi_2020}. However, their application to analyzing self-generated documents in RAG remains underexplored. Our work addresses this by introducing a taxonomy of Self-Docs grounded in Systemic Functional Linguistics (SFL)~\cite{Halliday1989LanguageCA}, optimizing document generation for factual accuracy and engagement in RAG systems.

\paragraph{Hybrid Strategies for Knowledge Integration}
Hybrid approaches, which combine retrieved and generated content, have shown promise in knowledge intensive tasks \citep{abdallah2024generatorretrievergeneratorapproachopendomainquestion, zhang2023merginggeneratedretrievedknowledge,yu2023generateretrievelargelanguage,shi-etal-2024-generate}. However, these methods often overlook how different types of generated documents affect performance. We propose a novel framework that classifies Self-Docs based on their communicative roles using SFL~\cite{Halliday1989LanguageCA}, allowing for more effective integration of generated and retrieved content, resulting in notable performance gains in knowledge-driven NLP tasks.

\section{Conclusion}
This study demonstrates that the type of self-generated document (Self-Docs) notably influences the effectiveness of Retrieval-Augmented Generation (RAG) systems. Tailoring Self-Docs to specific tasks leads to significant performance improvements across diverse knowledge-intensive tasks (RQ1 \& RQ2), and strategically integrating Self-Docs with external knowledge sources such as Wikipedia yields even greater gains (RQ3). In essence, scaling model capacity, adjusting document attributes (quantity, structure, tone, and granularity) to the task, and harmonizing these documents with external sources collectively maximize RAG performance.
\section*{Limitations}
This study primarily focuses on the Qwen2.5-32B-Instruct model due to computational constraints. Exploring the effectiveness of Self-Docs with larger models like GPT-4, especially in specialized domains requiring extensive knowledge, remains a promising direction for future work.

\section*{Ethical Considerations}
In this research, we evaluate the role of Self-Docs by models for enhancing the RAG system. One concern that arises is the inherent uncertainty regarding the correctness of the self-generated content. This issue is also prevalent in various methods such as self-generated data augmentation, self-reflection, and so on, as models may produce inaccurate or misleading information. While we have ensured the quality of the style of Self-Docs through human evaluation, their factual accuracy cannot be fully guaranteed.  

\section*{Acknowledgments}
This work was supported by Beijing Science and Technology Program (Z231100007423011) and Key Laboratory of Science, Technology and Standard in Press Industry (Key Laboratory of Intelligent Press Media Technology). We appreciate the anonymous reviewers for their helpful comments. Xiaojun Wan is the corresponding author.

\bibliography{main}
\appendix

\section{Tasks and Datasets}
\label{sec:dataset_details}

\paragraph{Open-Domain Question Answering}  
We evaluate open-domain QA on the TriviaQA (TQA)~\cite{joshi-etal-2017-triviaqa} dataset. TQA contains questions $q$ alongside answers $o$, extracted from supporting Wikipedia documents $P$. We evaluate the dataset using the Exact Match (EM) metric, following the framework used by~\citet{NEURIPS2020_6b493230}. 

\paragraph{Multi-Hop Question Answering}  
For multi-hop reasoning, we use the HotpotQA dataset~\cite{yang2018hotpotqadatasetdiverseexplainable}, which requires reasoning across multiple passages $P$ to arrive at a correct answer. HotpotQA includes over 113K question-answer pairs, many of which necessitate reasoning beyond a single source document. This task is evaluated using the F$_1$ score, as proposed by~\citet{yang2018hotpotqadatasetdiverseexplainable}, due to the abstractive nature of many responses.

\paragraph{Fact Verification}  
We assess fact verification capabilities using the FEVER dataset~\cite{thorne2018feverlargescaledatasetfact}, part of the KILT benchmark~\cite{petroni2021kiltbenchmarkknowledgeintensive}. Each sample consists of a claim $q$ that is either supported or refuted by Wikipedia passages. Claims are labeled as "SUPPORTS" or "REFUTES" based on whether they align with or contradict the factual evidence. We report performance using the Exact Match (EM) metric, consistent with prior work~\cite{thorne2018feverlargescaledatasetfact}.

\paragraph{Long-Form Question Answering}  
For long-form QA, we use the ELI5 dataset~\cite{fan2019eli5longformquestion}, which contains 270K threads from the Reddit forum “Explain Like I’m Five.” These questions require elaborate, multi-sentence answers, often demanding a higher level of abstraction and reasoning. Given the complexity of the responses, we evaluate performance using the F$_1$ metric, focusing on the generative short-answer setting, where models must provide concise yet informative responses.

\section{Implementation Details}  
\label{sec:implement_details}  

This study excludes the Natural Questions (NQ) dataset due to multiple issues identified in prior research. A significant portion of NQ questions (approximately 16.5\%) are time-sensitive, leading to evaluation inconsistencies when temporal context is not explicitly considered \citep{zhang2021situatedqa,yu2023generateretrievelargelanguage}. Additionally, discrepancies in the Wikipedia versions used for retrieval result in substantial performance variability \citep{izacard2022few,yu2023generateretrievelargelanguage}. Furthermore, error analysis from \textit{GenRead} highlighted data quality concerns in NQ, including annotation errors, incomplete answers, and retrieval inconsistencies \citep{yu2023generateretrievelargelanguage}. These challenges undermine the rigor and accuracy of assessments, and thus NQ is excluded from our study.

Our implementation follows a two-step workflow: generating diverse Wiki context shots and transforming them into target types to produce Self-Docs of various styles. This approach ensures that the generated documents align closely with the desired types, enabling the model to provide tailored, high-quality contexts for retrieval-augmented generation (RAG) systems. For shot generation, we utilize the \texttt{bge-large-en-v1.5} model \citep{bge_embedding} to embed both questions and documents, following the methodology outlined in \textit{GenRead} \citep{yu2023generateretrievelargelanguage}. This process produces a set of diverse "shots," each consisting of a query paired with its corresponding top-ranked Wikipedia context. However, the factual, encyclopedic style of Wikipedia articles often diverges from the desired Self-Docs types.

To generate Self-Docs with diverse styles, we extend the process by employing GPT-4o for style transformation. Retrieved Wikipedia contexts are transformed into target styles using task-specific prompts.\footnote{We use contriever \citep{izacard2021contriever} to retrieve wiki documents.} The style transformation function modifies documents to align with predefined formats (e.g., conversational, fine-grained), resulting in style-adapted outputs under a zero-shot setting. These transformed examples are then used to guide our backbone model, Qwen2.5-32B-Instruct, to generate Self-Docs. The temperature parameter is set to 0.95 during document generation to encourage diversity, while greedy decoding is used for question answering inference (temperature = 0) to ensure deterministic outputs and reproducibility. We also experiment with varying the number of documents (\( n = \{1, 3, 5, 10\} \)) to evaluate the impact of document count on system performance.

A manual evaluation of 120 sampled Self-Docs revealed that 94.2\% of the generated documents matched the target types in tone, granularity, and structure, demonstrating the efficacy of the style transformation process. Details of this evaluation are presented in Appendix~\ref{sec:manual_analysis}, while the GPT-4o style transformation and Qwen2.5-32B-Instruct prompts are detailed in Appendices~\ref{sec:shot_gen_prompt} and \ref{sec:prompt_docs_gen}, respectively. This setup ensures consistent, high-quality document generation tailored to the specific demands of the RAG framework.  
\section{Manual Evaluation}
\label{sec:manual_analysis}

To assess whether the self-generated documents (Self-Docs) align with their expected types, we conducted a manual evaluation of 120 document samples from the TQA dataset, covering the eight Self-Docs types outlined in Table~\ref{tab:document_types}. We selected 15 samples from each Self-Docs type and manually evaluated how well each document matched its designated interpersonal, ideational, and textual metafunctions. The success rates are presented in Table~\ref{tab:manual_analysis}.

\begin{table}[h]
\centering
\small
\setlength{\tabcolsep}{4pt}
\renewcommand{\arraystretch}{1.25}
\resizebox{\columnwidth}{!}{
\begin{tabular}{cccccc}
\toprule
\textbf{Type} & \textbf{Interpersonal} & \textbf{Ideational} & \textbf{Textual} & \textbf{Success Rate (\%)}\\
\midrule
AFS & Authoritative & Fine-Grain & Structured  & 86.7 \\
AFU & Authoritative & Fine-Grain & Unstructured & 86.7 \\
ACS & Authoritative & Coarse-Grain & Structured & 93.3 \\
ACU & Authoritative & Coarse-Grain & Unstructured & 100.0 \\
CFS & Conversational & Fine-Grain & Structured & 93.3 \\
CFU & Conversational & Fine-Grain & Unstructured & 93.3 \\
CCS & Conversational & Coarse-Grain & Structured & 100.0 \\
CCU & Conversational & Coarse-Grain & Unstructured & 100.0 \\
\bottomrule
\end{tabular}}
\caption{Success rates of self-generated document types classified based on the three SFL metafunctions: interpersonal, ideational, and textual.}
\label{tab:manual_analysis}
\end{table}

Each document was evaluated for its consistency with the SFL metafunctions, and the results indicate high success rates across all types, with some achieving perfect alignment. These results underscore the reliability of the Self-Docs taxonomy in producing documents that align with their expected communicative roles.

\section{Prompts}
We provide detailed prompts used throughout the experiment. The prompt for shot generation can be found in Section~\ref{sec:shot_gen_prompt}, while the prompt for generating context documents is presented in Section~\ref{sec:prompt_docs_gen}. Finally, the prompt used for the QA tasks is outlined in Section~\ref{sec:qa_prompt}. 
\label{sec:prompt}
\subsection{Prompt for shot generation}
\label{sec:shot_gen_prompt}
\begin{prompt}[title={Prompt \thetcbcounter: TQA, HotpotQA, Eli5 - AFS}]
Given the context: \{context\} and the query: \{query\}, transform the context into an authoritative, fine-grain, structured document.\\ The tone should be authoritative, maintaining a formal and objective style.\\ The content should be fine-grain, providing detailed explanations and in-depth analysis of each aspect.\\ The structure should be clear with proper headings and sections to logically organize the content.\\
\end{prompt}

\begin{prompt}[title={Prompt \thetcbcounter: TQA, HotpotQA, Eli5 - AFU}]
Given the context: \{context\} and the query: \{query\}, transform the context into an authoritative, fine-grain, unstructured document.\\The tone should be authoritative, maintaining a formal and objective style.\\The content should be fine-grain, providing detailed explanations and in-depth analysis of each aspect.\\The structure should be free-flowing, allowing ideas to develop organically without strict headings.
\end{prompt}

\begin{prompt}[title={Prompt \thetcbcounter: TQA, HotpotQA, Eli5-ACS}]
Given the context: \{context\} and the query: \{query\}, transform the context into an authoritative, coarse-grain, structured document.\\The tone should be authoritative, maintaining a formal and objective style.\\The content should be coarse-grain, offering a broad overview and key points without deep details.\\The structure should be clear with proper headings and sections to logically organize the content.
\end{prompt}    

\begin{prompt}[title={Prompt \thetcbcounter: TQA, HotpotQA, Eli5 - ACU}]
Given the context: \{context\} and the query: \{query\}, transform the context into an authoritative, coarse-grain, unstructured document.\\The tone should be authoritative, maintaining a formal and objective style.\\The content should be coarse-grain, offering a broad overview and key points without deep details.\\The structure should be free-flowing, allowing ideas to develop organically without strict headings.
\end{prompt}    

\begin{prompt}[title={Prompt \thetcbcounter: TQA, HotpotQA, Eli5 - CFS}]
Given the context: \{context\} and the query: \{query\}, transform the context into a conversational, fine-grain, structured document.\\The tone should be conversational, using an informal and approachable style.\\The content should be fine-grain, providing detailed explanations and in-depth analysis of each aspect.\\The structure should be clear with proper headings and sections to logically organize the content.
\end{prompt}    

\begin{prompt}[title={Prompt \thetcbcounter: TQA, HotpotQA, Eli5 - CFU}]
Given the context: \{context\} and the query: \{query\}, transform the context into a conversational, fine-grain, unstructured document.\\The tone should be conversational, using an informal and approachable style.\\The content should be fine-grain, providing detailed explanations and in-depth analysis of each aspect.\\The structure should be free-flowing, allowing ideas to develop organically without strict headings.
\end{prompt}    

\begin{prompt}[title={Prompt \thetcbcounter: TQA, HotpotQA, Eli5 - CCS}]
Given the context: \{context\} and the query: \{query\}, transform the context into a conversational, coarse-grain, structured document.\\The tone should be conversational, using an informal and approachable style.\\The content should be coarse-grain, offering a broad overview and key points without deep details.\\The structure should be clear with proper headings and sections to logically organize the content.
\end{prompt}    

\begin{prompt}[title={Prompt \thetcbcounter: TQA, HotpotQA, Eli5 - CCU}]
Given the context: \{context\} and the query: \{query\}, transform the context into a conversational, coarse-grain, unstructured document.\\The tone should be conversational, using an informal and approachable style.\\The content should be coarse-grain, offering a broad overview and key points without deep details.\\The structure should be free-flowing, allowing ideas to develop organically without strict headings.
\end{prompt}    

\begin{prompt}[title={Prompt \thetcbcounter: Fever - AFS}]
Given the context: \{context\} and the claim: \{query\}, transform the context into an authoritative, fine-grain, structured document to support or refute the claim.\\The tone should be authoritative, maintaining a formal and objective style.\\The content should be fine-grain, providing detailed explanations and in-depth analysis of each aspect.\\The structure should be clear with proper headings and sections to logically organize the content.
\end{prompt}

\begin{prompt}[title={Prompt \thetcbcounter: Fever - AFU}]
Given the context: \{context\} and the claim: \{query\}, transform the context into an authoritative, fine-grain, unstructured document to support or refute the claim.\\The tone should be authoritative, maintaining a formal and objective style.\\The content should be fine-grain, providing detailed explanations and in-depth analysis of each aspect.\\The structure should be free-flowing, allowing ideas to develop organically without strict headings.
\end{prompt}

\begin{prompt}[title={Prompt \thetcbcounter: Fever - ACS}]
Given the context: \{context\} and the claim: \{query\}, transform the context into an authoritative, coarse-grain, structured document to support or refute the claim.\\The tone should be authoritative, maintaining a formal and objective style.\\The content should be coarse-grain, offering a broad overview and key points without deep details.\\The structure should be clear with proper headings and sections to logically organize the content.
\end{prompt}

\begin{prompt}[title={Prompt \thetcbcounter:  Fever - ACU}]
Given the context: \{context\} and the claim: \{query\}, transform the context into an authoritative, coarse-grain, unstructured document to support or refute the claim.\\The tone should be authoritative, maintaining a formal and objective style.\\The content should be coarse-grain, offering a broad overview and key points without deep details.\\The structure should be free-flowing, allowing ideas to develop organically without strict headings.
\end{prompt}

\begin{prompt}[title={Prompt \thetcbcounter: Fever - CFS}]
Given the context: \{context\} and the claim: \{query\}, transform the context into a conversational, fine-grain, structured document to support or refute the claim.\\The tone should be conversational, using an informal and approachable style.\\The content should be fine-grain, providing detailed explanations and in-depth analysis of each aspect.\\The structure should be clear with proper headings and sections to logically organize the content.
\end{prompt}

\begin{prompt}[title={Prompt \thetcbcounter:  Fever - CFU}]
Given the context: \{context\} and the claim: \{query\}, transform the context into a conversational, fine-grain, unstructured document to support or refute the claim.\\The tone should be conversational, using an informal and approachable style.\\The content should be fine-grain, providing detailed explanations and in-depth analysis of each aspect.\\The structure should be free-flowing, allowing ideas to develop organically without strict headings.
\end{prompt}

\begin{prompt}[title={Prompt \thetcbcounter: Fever - CCS}]
Given the context: \{context\} and the claim: \{query\}, transform the context into a conversational, coarse-grain, structured document to support or refute the claim.\\The tone should be conversational, using an informal and approachable style.\\The content should be coarse-grain, offering a broad overview and key points without deep details.\\The structure should be clear with proper headings and sections to logically organize the content.
\end{prompt}

\begin{prompt}[title={Prompt \thetcbcounter: Fever - CCU}]
Given the context: \{context\} and the claim: \{query\}, transform the context into a conversational, coarse-grain, unstructured document to support or refute the claim.\\The tone should be conversational, using an informal and approachable style.\\The content should be coarse-grain, offering a broad overview and key points without deep details.\\The structure should be free-flowing, allowing ideas to develop organically without strict headings.
\end{prompt}

\subsection{Prompt for document generation}
\label{sec:prompt_docs_gen}

\begin{prompt}[title={Prompt \thetcbcounter: TQA, HotpotQA, Eli5 - AFS}]
System: You are tasked with generating a detailed and authoritative response. Maintain a formal and objective tone while providing in-depth analysis. The document should have a structured format with clear headings and sections that logically organize the content, offering precise explanations of each aspect.\\ \\
User: Given the query: \{query\}, generate an authoritative, fine-grain, structured document.\\ The tone should be authoritative, maintaining a formal and objective style.\\ The content should be fine-grain, providing detailed explanations and in-depth analysis of each aspect.\\ The structure should be clear with proper headings and sections to logically organize the content.
\end{prompt}

\begin{prompt}[title={Prompt \thetcbcounter: TQA, HotpotQA, Eli5 - AFU}]
System: You are tasked with generating a detailed and authoritative response in a free-flowing format. The tone should remain formal and objective, while the content offers in-depth analysis. The document should allow ideas to develop naturally without the constraint of rigid headings, encouraging an organic flow of detailed information.\\ \\
User: Given the query: \{query\}, generate an authoritative, coarse-grain, structured document.\\ The tone should be authoritative, maintaining a formal and objective style.\\ The content should be coarse-grain, offering a broad overview and key points without deep details.\\ The structure should be clear with proper headings and sections to logically organize the content.
\end{prompt}

\begin{prompt}[title={Prompt \thetcbcounter: TQA, HotpotQA, Eli5 - ACU}]
System: You are tasked with generating an authoritative, high-level overview in a free-flowing format. The tone should remain formal and objective, while the content offers broad explanations without going into detail. The document should allow ideas to develop organically without strict structure, focusing on the overall flow of the argument.\\ \\
User: Given the query: \{query\}, generate an authoritative, coarse-grain, unstructured document.\\ The tone should be authoritative, maintaining a formal and objective style.\\ The content should be coarse-grain, offering a broad overview and key points without deep details.\\ The structure should be free-flowing, allowing ideas to develop organically without strict headings.
\end{prompt}

\begin{prompt}[title={Prompt \thetcbcounter: TQA, HotpotQA, Eli5 - CFS}]
System: You are tasked with generating a detailed and engaging document. Use a conversational tone that is approachable and informal while maintaining a structured format. The content should provide in-depth explanations, with clear headings and sections that logically guide the reader through the document.\\ \\
User: Given the query: \{query\}, generate a conversational, fine-grain, structured document.\\The tone should be conversational, using an informal and approachable style.\\The content should be fine-grain, providing detailed explanations and in-depth analysis of each aspect.\\The structure should be clear with proper headings and sections to logically organize the content.
\end{prompt}

\begin{prompt}[title={Prompt \thetcbcounter: TQA, HotpotQA, Eli5 - CFU}]
System: You are tasked with generating a detailed and engaging response in a conversational tone. Use an informal and approachable style while allowing the document to flow naturally without strict headings or sections. Provide in-depth explanations while maintaining a free-flowing, organic narrative structure.\\ \\
User: Given the query: \{query\}, generate a conversational, fine-grain, unstructured document.\\The tone should be conversational, using an informal and approachable style.\\The content should be fine-grain, providing detailed explanations and in-depth analysis of each aspect.\\The structure should be free-flowing, allowing ideas to develop organically without strict headings.
\end{prompt}

\begin{prompt}[title={Prompt \thetcbcounter: TQA, HotpotQA, Eli5 - CCS}]
System: You are tasked with generating a conversational, high-level overview. Maintain an informal and approachable tone while organizing the content into clear headings and sections. The content should offer a broad overview, summarizing key points without going into much detail.\\ \\
User: Given the query: \{query\}, generate a conversational, coarse-grain, structured document.\\The tone should be conversational, using an informal and approachable style.\\The content should be coarse-grain, offering a broad overview and key points without deep details.\\The structure should be clear with proper headings and sections to logically organize the content.
\end{prompt}

\begin{prompt}[title={Prompt \thetcbcounter: TQA, HotpotQA, Eli5 - CCU}]
System: You are tasked with generating a conversational, broad overview in a free-flowing format. Use an informal and approachable tone while allowing the content to develop organically without a structured format. The document should provide key points and high-level explanations without strict headings or detailed sections.\\ \\
User: Given the query: \{query\}, generate a conversational, coarse-grain, unstructured document.\\The tone should be conversational, using an informal and approachable style.\\The content should be coarse-grain, offering a broad overview and key points without deep details.\\The structure should be free-flowing, allowing ideas to develop organically without strict headings.
\end{prompt}

\begin{prompt}[title={Prompt \thetcbcounter: Fever - AFS}]
System: You are tasked with generating a detailed, authoritative analysis to support or refute a claim. Maintain a formal and objective tone, providing in-depth analysis and explanations of each aspect. The document should have a structured format with clear headings and sections, logically organizing the reasoning.\\ \\
User: Given the claim: \{claim\}, generate an authoritative, fine-grain, structured document to support or refute the claim.\\The tone should be authoritative, maintaining a formal and objective style.\\The content should be fine-grain, providing detailed explanations and in-depth analysis of each aspect.\\The structure should be clear with proper headings and sections to logically organize the content.
\end{prompt}

\begin{prompt}[title={Prompt \thetcbcounter: Fever - AFU}]
System: You are tasked with generating a detailed, authoritative analysis in a free-flowing format to support or refute a claim. Maintain a formal and objective tone, while allowing the document to develop naturally without rigid headings. The content should provide in-depth explanations and analysis, letting ideas unfold organically.\\ \\
User: Given the claim: \{claim\}, generate an authoritative, fine-grain, unstructured document to support or refute the claim.\\The tone should be authoritative, maintaining a formal and objective style.\\The content should be fine-grain, providing detailed explanations and in-depth analysis of each aspect.\\The structure should be free-flowing, allowing ideas to develop organically without strict headings.
\end{prompt}

\begin{prompt}[title={Prompt \thetcbcounter: Fever - ACS}]
System: You are tasked with generating a broad, authoritative analysis to support or refute a claim. Maintain a formal and objective tone, summarizing the key points without deep details. The document should have clear headings and sections to logically organize the argument.\\ \\
User: Given the claim: \{claim\}, generate an authoritative, coarse-grain, structured document to support or refute the claim.\\The tone should be authoritative, maintaining a formal and objective style.\\The content should be coarse-grain, offering a broad overview and key points without deep details.\\The structure should be clear with proper headings and sections to logically organize the content.
\end{prompt}

\begin{prompt}[title={Prompt \thetcbcounter: Fever - ACU}]
System: You are tasked with generating a high-level, authoritative analysis in a free-flowing format to support or refute a claim. Maintain a formal and objective tone while allowing the content to develop organically without structured sections. Summarize key points without delving into too much detail.\\ \\
User: Given the claim: \{claim\}, generate an authoritative, coarse-grain, unstructured document to support or refute the claim.\\The tone should be authoritative, maintaining a formal and objective style.\\The content should be coarse-grain, offering a broad overview and key points without deep details.\\The structure should be free-flowing, allowing ideas to develop organically without strict headings.
\end{prompt}

\begin{prompt}[title={Prompt \thetcbcounter: Fever - CFS}]
System: You are tasked with generating a detailed, conversational analysis to support or refute a claim. Use an informal and approachable tone while maintaining a structured format with clear headings and sections. Provide in-depth explanations and a logical argument to support or refute the claim.\\ \\
User: Given the claim: \{claim\}, generate a conversational, fine-grain, structured document to support or refute the claim.\\The tone should be conversational, using an informal and approachable style.\\The content should be fine-grain, providing detailed explanations and in-depth analysis of each aspect.\\The structure should be clear with proper headings and sections to logically organize the content.
\end{prompt}

\begin{prompt}[title={Prompt \thetcbcounter: Fever - CFU}]
System: You are tasked with generating a detailed, conversational analysis to support or refute a claim in a free-flowing format. Use an informal and approachable tone while allowing the content to develop naturally without rigid structure. Provide in-depth explanations and analysis while letting ideas flow organically.\\ \\
User: Given the claim: \{claim\}, generate a conversational, fine-grain, unstructured document to support or refute the claim.\\The tone should be conversational, using an informal and approachable style.\\The content should be fine-grain, providing detailed explanations and in-depth analysis of each aspect.\\The structure should be free-flowing, allowing ideas to develop organically without strict headings.
\end{prompt}

\begin{prompt}[title={Prompt \thetcbcounter: Fever - CCS}]
System: You are tasked with generating a conversational, high-level analysis to support or refute a claim. Use an informal and approachable tone while organizing the content into clear headings and sections. Summarize key points without going into much detail, maintaining a broad overview.\\ \\
User: Given the claim: \{claim\}, generate a conversational, coarse-grain, structured document to support or refute the claim.\\The tone should be conversational, using an informal and approachable style.\\The content should be coarse-grain, offering a broad overview and key points without deep details.\\The structure should be clear with proper headings and sections to logically organize the content.
\end{prompt}

\begin{prompt}[title={Prompt \thetcbcounter: TQA, HotpotQA, Eli5 - GenRead}]
Provide a background document from Wikipedia to answer the given question. \\ \\
\{query\} \\ \\ 
\{wiki top1 context\} \\ \\ \\ \\
Provide a background document from Wikipedia to answer the given question. \\ \\
\{query\} \\ \\
\{wiki top1 context\} \\ \\ \\ \\
Provide a background document from Wikipedia to answer the given question. \\ \\
\{query\} \\ \\
\{wiki top1 context\} \\ \\ \\ \\
Provide a background document from Wikipedia to answer the given question. \\ \\
\{query\} \\ \\
\{wiki top1 context\} \\ \\ \\ \\
Provide a background document from Wikipedia to answer the given question. \\ \\
\{query\} \\ \\
\{wiki top1 context\} \\ \\ \\ \\
Provide a background document from Wikipedia to answer the given question. \\ \\
\{query\} \\ \\
\end{prompt}

\begin{prompt}[title={Prompt \thetcbcounter: Fever - GenRead}]
Generate a background document from Wikipedia to support or refute the statement. \\ \\
Statement: \{query\} \\ \\
\{wiki top1 context\} \\ \\ \\ \\
Generate a background document from Wikipedia to support or refute the statement. \\ \\
Statement: \{query\} \\ \\
\{wiki top1 context\} \\ \\ \\ \\
Generate a background document from Wikipedia to support or refute the statement. \\ \\
Statement: \{query\} \\ \\
\{wiki top1 context\} \\ \\ \\ \\
Generate a background document from Wikipedia to support or refute the statement. \\ \\
Statement: \{query\} \\ \\
\{wiki top1 context\} \\ \\ \\ \\
Generate a background document from Wikipedia to support or refute the statement. \\ \\
Statement: \{query\} \\ \\
\{wiki top1 context\} \\ \\ \\ \\
Generate a background document from Wikipedia to support or refute the statement. \\ \\
Statement: \{query\} \\ \\
\end{prompt}

\begin{prompt}[title={Prompt \thetcbcounter: Fever - CCU}]
System: You are tasked with generating a conversational, high-level analysis in a free-flowing format to support or refute a claim. Use an informal and approachable tone, allowing ideas to develop organically without rigid structure. Summarize key points without going into too much detail.\\ \\
User: Given the claim: \{claim\}, generate a conversational, coarse-grain, unstructured document to support or refute the claim.\\The tone should be conversational, using an informal and approachable style.\\The content should be coarse-grain, offering a broad overview and key points without deep details.\\The structure should be free-flowing, allowing ideas to develop organically without strict headings.
\end{prompt}

\subsection{Prompt for question answering}
\label{sec:qa_prompt}
\begin{prompt}[title={Prompt \thetcbcounter: TQA, HotpotQA - No RAG}]
System: This is a chat between a user and an artificial intelligence assistant. The assistant gives helpful, detailed, and polite answers to the user's questions.\\ \\
\{query\}. Answer the above question. Your output should strictly be one entity.
\end{prompt}

\begin{prompt}[title={Prompt \thetcbcounter: TQA, HotpotQA - RAG}]
System: This is a chat between a user and an artificial intelligence assistant. The assistant gives helpful, detailed, and polite answers to the user's questions based on the context. The assistant should also indicate when the answer cannot be found in the context.\\ \\
\{background\} User: \{query\}. Answer the above question. Your output should strictly be one entity.
\end{prompt}

\begin{prompt}[title={Prompt \thetcbcounter: Fever - No RAG}]
System: This is a chat between a user and an artificial intelligence assistant. The assistant gives helpful, detailed, and polite answers to the user's questions.\\ \\
Answer the following question with True or False. Is the claim \{claim\} correct? Your output should be either True or False.
\end{prompt}

\begin{prompt}[title={Prompt \thetcbcounter: Fever - RAG}]
System: This is a chat between a user and an artificial intelligence assistant. The assistant gives helpful, detailed, and polite answers to the user's questions based on the context. The assistant should also indicate when the answer cannot be found in the context.\\ \\
\{background\} User: Answer the following question with True or False. Is the claim \{claim\} correct? Your output should be either True or False.
\end{prompt}

\begin{prompt}[title={Prompt \thetcbcounter: Eli5 - No RAG}]
System: This is a chat between a user and an artificial intelligence assistant. The assistant gives helpful, detailed, and polite answers to the user's questions.\\ \\
\{query\}
\end{prompt}

\begin{prompt}[title={Prompt \thetcbcounter: Eli5 - RAG}]
System: This is a chat between a user and an artificial intelligence assistant. The assistant gives helpful, detailed, and polite answers to the user's questions based on the context.\\ \\
\{background\} User: \{query\}
\end{prompt}

\section{Pseudocode}

\begin{algorithm}
    \caption{Mix with Style Transformation Strategy}
    \label{alg:style_transform_mix}
    \renewcommand{\algorithmicrequire}{\textbf{Input:}}
    \renewcommand{\algorithmicensure}{\textbf{Output:}}
    
    \begin{algorithmic}[1]
        \REQUIRE Set of ranked Wiki documents $W$, set of unordered self-generated documents (Self-Docs) $S$ from taxonomy category $T$, number of top Wiki documents $k$, number of Self-Docs $n$, transformation function $T(W_i)$
        \ENSURE Mixed document set $\mathcal{M}$

        \STATE Initialize empty list $\mathcal{M}$ to store the mixed document set
        
        \STATE \textbf{Select Top Wiki documents:}
        \FOR{each $W_i$ in the top $k$ Wiki documents from $W$}
            \STATE Apply style transformation: $T(W_i)$
            \STATE Add transformed document $T(W_i)$ to $\mathcal{M}$
        \ENDFOR
        
        \STATE \textbf{Select Self-Docs from Taxonomy Category:}
        \STATE Select the first $n$ Self-Docs from $S$, where $S$ belongs to a specific category $T$ in the Self-Docs taxonomy (e.g., \textit{Authoritative, Conversational, etc.})
        \STATE Add selected Self-Docs to $\mathcal{M}$
        
        \STATE \textbf{Final Mixed Document:}
        \[
        D_{\text{Mixed}} = \text{Concat}\left(\bigcup_{i=1}^{k} T(W_i), \bigcup_{j=1}^{n} S_j\right)
        \] 
        
        \RETURN $\mathcal{M}$
    
    \end{algorithmic}
\end{algorithm}

\begin{algorithm}
    \caption{Direct Mix Strategy for Document Mixing}
    \label{alg:direct_mix}
    \renewcommand{\algorithmicrequire}{\textbf{Input:}}
    \renewcommand{\algorithmicensure}{\textbf{Output:}}
    
    \begin{algorithmic}[1]
        \REQUIRE Set of ranked Wiki documents $W$, set of unordered self-generated documents (Self-Docs) $S$ from taxonomy category $T$, number of top Wiki documents $k$, number of Self-Docs $n$
        \ENSURE Mixed document set $\mathcal{M}$
    
        \STATE Initialize empty list $\mathcal{M}$ to store the mixed document set
        
        \STATE \textbf{Select Top Wiki documents:}
        \STATE Retrieve and select the top $k$ Wiki documents from $W$ based on rank order
        \STATE Add selected Wiki documents to $\mathcal{M}$
        
        \STATE \textbf{Select Self-Docs from Taxonomy Category:}
        \STATE Select the first $n$ Self-Docs from $S$, where $S$ belongs to a specific category $T$ in the Self-Docs taxonomy (e.g., \textit{Authoritative-Coarse-Grained-Structured})
        \STATE Add selected Self-Docs to $\mathcal{M}$
        
        \STATE \textbf{Final Mixed Document:}
        \[
        D_{\text{Mixed}} = \text{Concat}(D_{\text{Self-Docs}}, D_{\text{Wiki}})
        \] 
        
        \RETURN $\mathcal{M}$
    
    \end{algorithmic}
\end{algorithm}

\begin{algorithm}
    \caption{Self-Docs Construction Pipeline}
    \label{alg:sgd_pipeline}
    \renewcommand{\algorithmicrequire}{\textbf{Input:}}
    \renewcommand{\algorithmicensure}{\textbf{Output:}}
    
    \begin{algorithmic}[1]
        \REQUIRE Set of queries $Q = \{q_1, q_2, \ldots, q_n\}$, top-1 retrieved wiki documents $D = \{d_1, d_2, \ldots, d_n\}$, number of clusters $k$, number of few-shot examples per cluster $m$, target style $T$ (optional)
        \ENSURE Generated Self-Docs $G = \bigcup_{t=1}^n G_t = \{g_{1,1}, \ldots, g_{n,k}\}$
        
        \STATE \textbf{Input and Clustering:}
        \STATE Compute embeddings for each $(q_i + d_i)$ pair
        \STATE Apply $k$-means clustering to group embeddings into $k$ clusters $C = \{C_1, \ldots, C_k\}$
        
        \STATE \textbf{Few-Shot Examples:}
        \FOR{each cluster $C_i$}
            \STATE Sample $m$ query-document pairs $\text{FewShot}_j = \langle \text{Input: } q_j, \text{Output: } d_j \rangle$
        \ENDFOR
        
        \STATE \textbf{Self-Docs Generation:}
        \FOR{each query $q_t \in Q$}
            \FOR{each cluster $C_i$}
                \STATE Generate document $g_{t,i} = \mathcal{L}(q_t \mid \text{FewShots}_i)$
            \ENDFOR
        \ENDFOR
        
        \STATE \textbf{Style Alignment (if target style $T$ is provided):}
        \FOR{each document $d_j$ in $\text{FewShot}_j$}
            \STATE Transform $d_j$ into target style: $t_j = \text{StyleTransform}(d_j, T)$
            \STATE Update few-shot examples: $\text{FewShot}_j = \langle \text{Input: } q_j, \text{Output: } t_j \rangle$
        \ENDFOR
        
        \STATE \textbf{Output:}
        \STATE Combine all generated documents $G_t = \{g_{t,1}, \ldots, g_{t,k}\}$ for each $q_t$
        \STATE Final Self-Docs: $G = \bigcup_{t=1}^n G_t$
        
        \RETURN $G$
    \end{algorithmic}
\end{algorithm}
\section{Detailed Results}
\label{sec:detailed_results}
\begin{table}[h]
\centering
\label{tab:performance_n_values}
\begin{tabular}{l|cccc}
\toprule
\textbf{Dataset} & \textbf{n=10} & \textbf{n=5} & \textbf{n=3} & \textbf{n=1} \\
\midrule
\textsc{TQA}      & 56.6  & 54.0  & 52.0  & 49.6  \\
\textsc{FEVER}    & 88.8  & 87.6  & 86.6  & 83.8  \\
\textsc{HotpotQA} & 37.2  & 36.0  & 36.1  & 34.4  \\
\textsc{ELI5}     & 24.4  & 24.7  & 25.1  & 26.0  \\
\bottomrule
\end{tabular}
\caption{Performance across various document numbers (\textit{n}) for different datasets. The results show how the number of self-generated documents impacts model performance, with $n$ values set to 10, 5, 3, and 1.}
\end{table}

\begin{table*}[h]
\centering
\resizebox{\textwidth}{!}{
\begin{tabular}{c|c|c|c|c}
\toprule
\textbf{Dataset} & \textbf{Qwen2.5-32B-Instruct} & \textbf{Qwen2.5-14B-Instruct} & \textbf{Qwen2.5-7B-Instruct} & \textbf{Qwen2.5-3B-Instruct} \\
\midrule
\textsc{TQA} & 56.6 & 55.0 & 45.2 & 33.0 \\
\textsc{FEVER} & 88.8 & 76.8 & 64.0 & 72.4 \\
\textsc{HotpotQA} & 37.2 & 38.3 & 32.9 & 25.1 \\
\textsc{ELI5} & 24.4 & 23.8 & 22.1 & 20.9 \\
\bottomrule
\end{tabular}
}
\caption{Performance metrics across different datasets and model sizes under the self-generated documents (Self-Docs) setting, with $n=10$ documents. Both the QA and context generation models are from the Qwen2.5 series, ranging from 3B to 32B parameters.}
\end{table*}

\begin{table*}[h]
\centering
\resizebox{\textwidth}{!}{
\begin{tabular}{c|c|c|c|c|c|c}
\toprule
\textbf{Info Source} & \textbf{Qwen2.5-32B-Instruct} & \textbf{Qwen2.5-14B-Instruct} & \textbf{Qwen2.5-7B-Instruct} & \textbf{Qwen2.5-3B-Instruct} & \textbf{Wiki} & \textbf{No-RAG} \\
\midrule
\textsc{TQA} & 56.6 & 53.2 & 45.8 & 39.6 & 52.6 & 52.0 \\
\textsc{FEVER} & 88.8 & 83.8 & 78.8 & 73.6 & 66.8 & 77.2 \\
\textsc{HotpotQA} & 37.2 & 39.1 & 33.0 & 26.0 & 36.3 & 32.9 \\
\textsc{ELI5} & 24.4 & 23.5 & 22.7 & 22.8 & 26.7 & 26.8 \\
\bottomrule
\end{tabular}
}
\caption{Performance comparison across different information sources (Qwen2.5 models of varying sizes, Wiki, and No-RAG) on multiple datasets. This table highlights the impact of model size and information source on retrieval-augmented tasks like TQA, FEVER, HotpotQA, and ELI5.}
\label{tab:figure3_performance}
\end{table*}


\begin{table*}[h]
\centering
\small
\resizebox{\textwidth}{!}{
\begin{tabular}{c|c|c|c|c|c|c|c|c}
\toprule
\textbf{Type} & \textbf{Interpersonal} & \textbf{Ideational} & \textbf{Textual} & \textbf{\textsc{TQA}} & \textbf{\textsc{HotpotQA}} & \textbf{\textsc{FEVER}} & \textbf{\textsc{ELI5}} & \textbf{Average} \\
\midrule
\textsc{AFS} & Authoritative & Fine-Grain & Structured  & 56.0\scriptsize{↑0.6} & 38.3\scriptsize{↑0.8} & 73.6\scriptsize{↓9.8} & 22.0\scriptsize{↑1.1} & 47.5\scriptsize{↓1.8} \\
\textsc{AFU} & Authoritative & Fine-Grain & Unstructured & 56.8\scriptsize{↓1.6} & 41.0\scriptsize{↑1.1} & 84.4\scriptsize{↓0.8} & 23.1\scriptsize{↑0.3} & 51.3\scriptsize{↓0.3} \\
\textsc{ACS} & Authoritative & Coarse-Grain & Structured & 55.8\scriptsize{↓1.0} & 39.5\scriptsize{↑0.8} & 84.8\scriptsize{↑1.2} & 22.7\scriptsize{↑0.7} & 50.7\scriptsize{↑0.4} \\
\textsc{ACU} & Authoritative & Coarse-Grain & Unstructured & 56.2\scriptsize{↓0.2} & 40.8\scriptsize{↑1.2} & \underline{86.6}\scriptsize{↑1.6} & 23.8\scriptsize{↑0.3} & 51.8\scriptsize{↑0.7} \\
\textsc{CFS} & Conversational & Fine-Grain & Structured & 56.4\scriptsize{↓1.0} & 39.0\scriptsize{↑2.6} & 78.2\scriptsize{↓3.6} & 24.8\scriptsize{↑0.8} & 49.6\scriptsize{↓0.3} \\
\textsc{CFU} & Conversational & Fine-Grain & Unstructured & 57.2\scriptsize{↓0.6} & 41.6\scriptsize{↑2.1} & 85.0\scriptsize{↑0.8} & 25.6\scriptsize{↑0.5} & 52.4\scriptsize{↑0.8} \\
\textsc{CCS} & Conversational & Coarse-Grain & Structured & 56.4\scriptsize{↑0.4} & 40.1\scriptsize{↑4.9} & 84.0\scriptsize{↑2.0} & 25.7\scriptsize{↑0.3} & 51.6\scriptsize{↑2.0} \\
\textsc{CCU} & Conversational & Coarse-Grain & Unstructured & \underline{57.4}\scriptsize{↑0.8} & \underline{\textbf{41.8}}\scriptsize{↑3.1} & 85.8\scriptsize{↑2.0} & \underline{26.5}\scriptsize{↑0.1} & \underline{\textbf{52.9}}\scriptsize{↑1.5} \\
\midrule
\textsc{Wiki} & Authoritative & Flexible & Unstructured & 52.6 & 36.3 & 66.8 & \textbf{26.7} & 45.6  \\
\textsc{GenRead mix} & Authoritative & Flexible & Unstructured & \textbf{57.6}  & 39.1  & \textbf{87.0}  & 24.7  & 52.1  \\
\bottomrule
\end{tabular}
}
\caption{Performance comparison across different generated document types combined with Wiki documents for various tasks. Superscript values indicate the difference from Table~\ref{tab:rq2_full}, with increases marked by an upward arrow (↑) and decreases by a downward arrow (↓). The best overall performance for each dataset is highlighted in \textbf{bold}, and the best performance within the category of Self-Docs with the same mix strategy is \underline{underlined}.}
\label{tab:rq3_full_direct_mix_with_raw_wiki}
\end{table*}

\begin{table*}[ht]
\vspace{-1mm}
\centering
    \begin{tabular}{c|c|cccc|c}
        \toprule
        \textbf{Meta Function} & \textbf{Dimension} & \textbf{\textsc{TQA}} & \textbf{\textsc{HotpotQA}} & \textbf{\textsc{FEVER}} & \textbf{\textsc{ELI5}} & \textbf{Average} \\
        \midrule
        \multirow{2}{*}[0em]{\textsc{Ideational}} 
                           & \textsc{Fine-Grain}      & \underline{56.6}\scriptsize{↓0.6} & 40.0\scriptsize{↑1.7} & 80.3\scriptsize{↓3.3} & 23.9\scriptsize{↑0.7} & 50.2\scriptsize{↓0.4} \\
                           & \textsc{Coarse-Grain}    & 56.4\scriptsize{±0.0} & \underline{40.6}\scriptsize{↑2.5} & \underline{85.3}\scriptsize{↑1.7} & \underline{24.7}\scriptsize{↑0.4} & \underline{51.8}\scriptsize{↑1.2} \\
        \midrule
        \multirow{2}{*}[0em]{\textsc{Interpersonal}} 
                           & \textsc{Authoritative}   & 56.2\scriptsize{↓0.6} & 39.9\scriptsize{↑1.0} & 82.4\scriptsize{↓1.9} & 22.9\scriptsize{↑0.6} & 50.4\scriptsize{↓0.2} \\
                           & \textsc{Conversational}  & \underline{56.8}\scriptsize{↓0.2} & \underline{40.6}\scriptsize{↑3.2} & \underline{83.2}\scriptsize{↑0.2} & \underline{25.6}\scriptsize{↑0.4} & \underline{51.6}\scriptsize{↑0.9} \\
        \midrule
        \multirow{2}{*}[0em]{\textsc{Textual}} 
                           & \textsc{Structured}      & 56.2\scriptsize{↓0.2} & 39.2\scriptsize{↑2.2} & 80.1\scriptsize{↓2.6} & 23.8\scriptsize{↑0.7} & 49.8\scriptsize{±0.0} \\
                           & \textsc{Unstructured}    & \underline{56.9}\scriptsize{↓0.4} & \underline{41.3}\scriptsize{↑1.9} & \underline{85.4}\scriptsize{↑0.8} & \underline{24.8}\scriptsize{↑0.3} & \underline{51.6}\scriptsize{↑0.2} \\
        \bottomrule
    \end{tabular}
\caption{Aggregated performance results across different dimensions of the metafunctions (Ideational, Interpersonal, and Textual) on various datasets. Superscript values indicate the difference from Table~\ref{tab:rq2_agg}, with increases marked by an upward arrow (↑) and decreases by a downward arrow (↓). The best performance within each metafunction dimension is \underline{underlined}.}
\vspace{-1mm}
\label{tab:rq3_agg_direct_mix_with_raw_wiki}
\end{table*}

\begin{table*}[h]
\centering
\resizebox{\textwidth}{!}{
\begin{tabular}{c|c|c|c|c|c|c|c|c}
\toprule
\textbf{Docs Type} & \textbf{Interpersonal} & \textbf{Ideational} & \textbf{Textual} & \textbf{\textsc{TQA}} & \textbf{\textsc{HotpotQA}} & \textbf{\textsc{FEVER}} & \textbf{\textsc{ELI5}} & \textbf{Average} \\
\midrule
\textsc{AFS} & Authoritative & Fine-Grain & Structured  & 56.6\scriptsize{↑1.2} & 37.5\scriptsize{±0.0} & 87.6\scriptsize{↑4.2} & 21.3\scriptsize{↑0.4} & 50.8\scriptsize{↑1.5} \\
\textsc{AFU} & Authoritative & Fine-Grain & Unstructured & \underline{\textbf{59.4}}\scriptsize{↑1.0} & \underline{\textbf{43.3}}\scriptsize{↑3.4} & 87.6\scriptsize{↑2.4} & 22.7\scriptsize{↓0.1} & 53.3\scriptsize{↑1.7} \\
\textsc{ACS} & Authoritative & Coarse-Grain & Structured & 56.8\scriptsize{±0.0} & 42.4\scriptsize{↑3.7} & 88.8\scriptsize{↑5.2} & 22.2\scriptsize{↑0.2} & 52.6\scriptsize{↑2.3} \\
\textsc{ACU} & Authoritative & Coarse-Grain & Unstructured & 57.0\scriptsize{↑0.6} & 41.9\scriptsize{↑2.3} & \underline{\textbf{90.4}}\scriptsize{↑5.4} & 23.3\scriptsize{↓0.2} & 53.2\scriptsize{↑2.1} \\
\textsc{CFS} & Conversational & Fine-Grain & Structured & 57.0\scriptsize{↓0.4} & 39.3\scriptsize{↑2.9} & 86.0\scriptsize{↑4.2} & 24.2\scriptsize{↑0.2} & 51.6\scriptsize{↑1.7} \\
\textsc{CFU} & Conversational & Fine-Grain & Unstructured & 58.0\scriptsize{↑0.2} & 41.0\scriptsize{↑1.5} & 86.4\scriptsize{↑2.2} & 25.3\scriptsize{↑0.2} & 52.7\scriptsize{↑1.1} \\
\textsc{CCS} & Conversational & Coarse-Grain & Structured & 57.4\scriptsize{↑1.4} & 41.8\scriptsize{↑6.6} & 86.6\scriptsize{↑4.6} & 25.6\scriptsize{↑0.2} & 52.9\scriptsize{↑3.3} \\
\textsc{CCU} & Conversational & Coarse-Grain & Unstructured & 57.6\scriptsize{↑1.0} & 43.0\scriptsize{↑4.3} & 87.4\scriptsize{↑3.6} & \underline{26.3}\scriptsize{↓0.1} & \underline{\textbf{53.6}}\scriptsize{↑2.2} \\
\midrule
\textsc{Wiki} & Authoritative & Flexible & Unstructured & 52.6  & 36.3  & 66.8  & \textbf{26.7}  & 45.6  \\
\textsc{GenRead mix} & Authoritative & Flexible & Unstructured & 57.6  & 39.1  & 87.0  & 24.7  & 52.1  \\
\bottomrule
\end{tabular}
}
\caption{Performance comparison across different generated document types combined with Wiki documents using mix with style transformation strategy for various tasks. Superscript values indicate the difference from Table~\ref{tab:rq2_full}, with increases marked by an upward arrow (↑) and decreases by a downward arrow (↓). The best overall performance for each dataset is highlighted in \textbf{bold}, and the best performance within the category of Self-Docs with the same mix strategy is \underline{underlined}.}
\label{tab:rq3_full_direct_mix_with_style_transformed_wiki}
\end{table*}

\begin{table*}[ht]
\vspace{-1mm}
\centering
    \begin{tabular}{c|c|cccc|c}
        \toprule
        \textbf{Meta Function} & \textbf{Dimension} & \textbf{\textsc{TQA}} & \textbf{\textsc{HotpotQA}} & \textbf{\textsc{FEVER}} & \textbf{\textsc{ELI5}} & \textbf{Average} \\
        \midrule
        \multirow{2}{*}[0em]{\textsc{Ideational}} 
                           & \textsc{Fine-Grain}      & \underline{57.8}\scriptsize{↑0.6} & 40.3\scriptsize{↑2.0} & 86.9\scriptsize{↑3.3} & 23.4\scriptsize{↑0.2} & 52.1\scriptsize{↑1.5} \\
                           & \textsc{Coarse-Grain}    & 57.2\scriptsize{↑0.8} & \underline{42.3}\scriptsize{↑4.2} & \underline{88.3}\scriptsize{↑4.7} & \underline{24.3}\scriptsize{±0.0} & \underline{53.0}\scriptsize{↑2.4} \\
        \midrule
        \multirow{2}{*}[0em]{\textsc{Interpersonal}} 
                           & \textsc{Authoritative}   & 56.5\scriptsize{↓0.3} & 40.3\scriptsize{↑1.4} & 84.2\scriptsize{↓0.1} & 23.2\scriptsize{↑0.9} & 51.0\scriptsize{↑0.4} \\
                           & \textsc{Conversational}  & \underline{57.5}\scriptsize{↑0.5} & \underline{41.3}\scriptsize{↑3.9} & \underline{86.6}\scriptsize{↑3.6} & \underline{25.3}\scriptsize{↑0.1} & \underline{52.7}\scriptsize{↑2.0} \\
        \midrule
        \multirow{2}{*}[0em]{\textsc{Textual}} 
                           & \textsc{Structured}      & \underline{57.0}\scriptsize{↑0.6} & 40.2\scriptsize{↑3.2} & \underline{87.2}\scriptsize{↑4.5} & 23.3\scriptsize{↑0.2} & \underline{51.9}\scriptsize{↑2.1} \\
                           & \textsc{Unstructured}    & 56.9\scriptsize{↓0.4} & \underline{41.1}\scriptsize{↑1.7} & 83.7\scriptsize{↓0.9} & \underline{24.9}\scriptsize{↑0.4} & 51.6\scriptsize{↑0.2} \\
        \bottomrule
    \end{tabular}

\caption{Aggregated performance results across different dimensions of the metafunctions (Ideational, Interpersonal, and Textual) on various datasets. Superscript values indicate the difference from Table~\ref{tab:rq2_agg}, with increases marked by an upward arrow (↑) and decreases by a downward arrow (↓). The best performance within each metafunction dimension is \underline{underlined}.}
\vspace{-1mm}
\label{tab:rq3_agg_direct_mix_with_style_transformed_wiki}
\end{table*}

\clearpage
\section{Case Study}
In this section, we present a case study leveraging the Qwen2.5-32B-Instruct model to assess its performance on self-generated documents (Self-Docs) across various document types, as defined by the full taxonomy in~\ref{sec:case_gen_docs}. This analysis explores how different Self-Docs types, classified according to interpersonal tone, ideational content, and textual structure, impact overall task performance. Additionally, we provide a case study evaluating the performance of GPT-4o-generated document shots in~\ref{sec:case_gen_shot}, demonstrating the effects of varying document styles on specific tasks. We also include an example case of QA with different styles of Self-Docs in~\ref{sec:QA_case}.

\subsection{Examples of Self-generated Document Shots}
\label{sec:case_gen_shot}

\begin{showcase}[title=Dataset: TQA]


|\classbg{Authoritative, Fine-Grain, Structured (AFS)}|
# The Sinatra Doctrine: A Foreign Policy Framework

## Introduction

The Sinatra Doctrine represents a pivotal shift in Soviet foreign policy, distinguishing itself by acknowledging the autonomy of Warsaw Pact countries in choosing their political paths. Named after Frank Sinatra's iconic song "My Way," this doctrine marked a significant departure from previous Soviet positions of influence over Eastern Europe.

## Historical Context

### The Soviet Union and Warsaw Pact Dynamics

The Warsaw Pact was a collective defense treaty established in 1955 among the Soviet Union and seven Eastern Bloc socialist republics during the Cold War. Traditionally, the Soviet Union exerted considerable influence over these states, often dictating their political and economic directions.

### Shift in Policy

The introduction of the Sinatra Doctrine emerged from evolving geopolitical realities in the late 20th century. As pressures for reform and democratization mounted across Eastern Europe, the Soviet leadership recognized the need to adapt its foreign policy stance to accommodate these changing dynamics.

## Formulation of the Sinatra Doctrine

The Sinatra Doctrine was articulated by Gennadi Gerasimov, a prominent Soviet government spokesperson, during an interview. He referenced Frank Sinatra's song "I Did It My Way" to illustrate the doctrine's essence: allowing each country to independently choose its path without imposed external influence.

### Key Statements

- **Gerasimov's Declaration:** Gerasimov affirmed, "We now have the Frank Sinatra doctrine. He has a song, 'I Did It My Way.' So every country decides on its own which road to take."
- **Acceptance of Political Diversity:** When queried about the potential rejection of communist parties within the Soviet bloc, Gerasimov stated, "That's for sure... political structures must be decided by the people who live there."

......

## Conclusion

The Sinatra Doctrine marked a significant transformation in Soviet foreign policy, emphasizing respect for national sovereignty and political self-determination. As an acknowledgment of evolving global trends, this doctrine played a crucial role in reshaping the political landscape of Eastern Europe during the late 20th century. Its legacy persists as a testament to the power of ideological evolution in response to geopolitical changes.

|\classbg{Authoritative, Fine-Grain, Unstructured (AFU)}|
In a significant ideological shift during the late 20th century, the Soviet Union introduced a notable transformation in its foreign policy approach, commonly referred to as the "Sinatra Doctrine." This strategic realignment was articulated by Soviet Foreign Ministry spokesman Gennadi Gerasimov in a pivotal moment of diplomatic evolution. The doctrine was symbolically named after the famed American singer Frank Sinatra, specifically invoking his song "My Way" to encapsulate the essence of this new policy.

The "Sinatra Doctrine" emerged as a clear departure from previous Soviet policies, which had traditionally exerted significant control over the internal and external affairs of the Warsaw Pact states. Historically, the Soviet Union maintained a policy of strict oversight, ensuring that the political and economic systems of Eastern Bloc countries were closely aligned with communist ideologies. However, Gerasimov's declaration marked a decisive shift towards acknowledging the sovereignty and autonomy of these nations.

During an interview, Gerasimov expounded on this new ideological perspective, emphasizing that the Soviet Union now recognized the principle of freedom of choice for all countries, explicitly including its Warsaw Pact allies. He articulated a vision where each nation possessed the authority to determine its own pathway, unhindered by external pressures. His statement, "We now have the Frank Sinatra doctrine. He has a song, 'I Did It My Way.' So every country decides on its own which road to take," served as a metaphorical representation of this newfound respect for national self-determination.

......

In essence, the "Sinatra Doctrine" laid the foundation for a transformative era in international relations. It embodied a philosophical pivot towards embracing pluralism, encouraging organic political developments, and enhancing diplomatic autonomy within the Eastern Bloc. Through this doctrine, the Soviet Union signaled a strategic retreat from its historically pervasive interventionist stance, thereby acknowledging and accommodating the complex realities of a rapidly changing world order.

|\classbg{Authoritative, Coarse-Grain, Structured (ACS)}|
# The Sinatra Doctrine: An Overview of a Decisive Policy Shift

## Introduction

The Sinatra Doctrine represents a significant shift in Soviet foreign policy during the late 20th century. It emphasizes the autonomy of individual countries, specifically those within the Warsaw Pact, to determine their own political paths independently of Soviet influence.

## Background

The doctrine emerged as a bold move acknowledging the changing geopolitical landscape. It marks a departure from previous Soviet policies that enforced strict adherence to communism across the Eastern Bloc.

## Key Principles

### Autonomy of Nations

- **Recognition of Freedom**: The doctrine asserts the right of each nation to choose its political structure and direction.
- **Rejection of Previous Controls**: It signifies a move away from the rigid enforcement of communist ideologies by the Soviet Union.

### Inspiration from Culture

- **Cultural Reference**: The doctrine derives its name from Frank Sinatra's famous song, "My Way," symbolizing the freedom of choice.
- **Philosophical Shift**: It underscores a philosophical shift towards respecting individual nation sovereignty.

......

## Conclusion

The Sinatra Doctrine stands as a pivotal moment in international relations, illustrating a refined approach to foreign policy grounded in the respect for national sovereignty. This realignment paved the way for significant political changes across Eastern Europe, marking the beginning of the end for Soviet dominance in the region.

|\classbg{Authoritative, Coarse-Grain, Unstructured (ACU)}|
In the context of evolving international relations during the late 20th century, a significant shift in Soviet foreign policy was articulated, highlighting a newfound recognition of national autonomy within the Eastern Bloc. This policy, informally dubbed the "Sinatra Doctrine," was articulated by a Soviet spokesperson. It was named after Frank Sinatra's iconic song "I Did It My Way," symbolizing the empowerment of each nation to independently determine their path.

The doctrine emerged as a pivotal change from previous policies, acknowledging the right of countries, particularly those within the Warsaw Pact, to exercise freedom of choice. This recognition extended to allowing them to decide their political structures without external interference, including the potential rejection of communist governments.

The authoritative declaration of this doctrine marked a departure from traditional Soviet influence, endorsing a more decentralized approach to governance among allied states. This strategy encouraged political determination by the populations residing within these nations, emphasizing self-determination as a cornerstone principle.

Through this declaration, the Soviet Union granted unprecedented autonomy to its neighbors, fostering a diplomatic environment where individual states were encouraged to pursue their unique political and social ideals. This transformation in policy underscored a broader movement towards recognizing the diversity of governance models within the international community, ultimately accepting that each country should establish its path in accordance with the will of its people.

|\classbg{Conversational, Fine-Grain, Structured (CFS)}|
# The Sinatra Doctrine: A Conversational Breakdown

## Introduction

Hey there! Let's dive into something intriguingly named the "Sinatra Doctrine." This wasn't some grand plan cooked up in a secret Kremlin meeting room, but rather a shift in thinking that signaled a big change in foreign policy. It's named after Frank Sinatra's classic song, "My Way." Intrigued? Let's break it down.

## What Is the Sinatra Doctrine?

### The Basics

So, what exactly are we talking about? The "Sinatra Doctrine" refers to a policy approach by the Soviet Union during the late 1980s. Picture this: instead of the Soviets dictating how the satellites in their orbit--like those in the Warsaw Pact--should operate, they started singing a new tune inspired by "I Did It My Way." Each country could now choose its own path without Moscow pulling the strings.

### Why Frank Sinatra?

You might be wondering why Frank Sinatra, of all people, was referenced for a Soviet policy. Well, it's because of his famous song "My Way." The idea was that just like in the song, every country could decide for itself what path to take--politically, economically, and socially.

## The Key Players

### Who Said What?

This fresh approach was put into simple words by a Soviet spokesman named Gerasimov. In an interview, he cheekily declared, "We now have the Frank Sinatra Doctrine." It was his way of saying that the Soviets were ready to let countries in the Eastern Bloc find their groove and do things their way.

### The Impact on Communism

Now, you might ask, did this mean the Soviets were okay with these countries rejecting communism? Gerasimov's answer was a straightforward "That's for sure." He emphasized that political decisions should be made by the people living in those countries--talking about major vibes of autonomy and independence!

......

## Conclusion

So, there you have it--the Sinatra Doctrine wasn't just about the Soviets letting go of control; it was about acknowledging the right of each nation to carve out its destiny. By evoking Sinatra's "My Way," it symbolized a significant turn in foreign policy, encouraging countries to embrace their sovereignty and decide their own future.

And that's it! Hope you enjoyed this deep dive into the surprising intersection of music and international policy.

|\classbg{Conversational, Fine-Grain, Unstructured (CFU)}|
Hey there! So, let's dive into this interesting bit about the so-called "Sinatra Doctrine," which is quite the shift from the old ways that were in place during the Soviet era.

Picture this: it's the late 1980s, and Mikhail Gorbachev comes into the scene with a fresh mindset. The world is changing, and the old Soviet grip on Eastern European countries is starting to loosen. Gorbachev's new man, Gerasimov, was kind of a big deal in this transition. He went on record saying that the Soviets were all about respecting the freedom of choice for all countries, which, at the time, was a pretty big deal.

It was in an interview where Gerasimov casually came up with the so-called "Sinatra Doctrine." You might be wondering why it's named after Sinatra. Well, here's the fun part: it was a nod to Frank Sinatra's famous song, "My Way." You know the one, right? It's all about doing things on your terms. Gerasimov said something like, "We're going with the Frank Sinatra Doctrine now. Every country gets to decide which road they want to take." This was sort of his cheeky way of saying that each country within the Soviet sphere could make its own choices.

Now, imagine someone asking Gerasimov, "So, does this really mean that if a country in the Soviet bloc decides to toss out its communist regime, you'd just accept that?" And Gerasimov was like, "Yeah, exactly. It's up to the people there to choose their political structures." This was a remarkable shift from the past when any deviation could have led to military interventions or other forms of pressure.

This whole "My Way" approach was radically different and paved the way for those countries in Eastern Europe to pursue their own paths without the fear of Soviet intervention. It was Gorbachev's way of signaling that the times were changing, and it opened the door for the fall of communism in several nations.

In essence, this doctrine wasn't some formal policy written in stone, but more a symbolic gesture indicating a real change in the political winds. It was about empowering nations to figure out their identities and futures independently. So, Gorbachev and his guys essentially said, "You do you, and we'll do our own thing." It was a critical moment leading to the massive political shifts in Europe during that era.

So, there you have it! The Sinatra Doctrine wasn't about what the Soviet Union would do but rather about what they wouldn't do--intervene in the choices of other countries. It was a big step away from the past, letting everyone sing their song "My Way."

|\classbg{Conversational, Coarse-Grain, Structured (CCS)}|
# The Sinatra Doctrine: A New Approach to Foreign Policy

## Introduction

Hey there! Ever wondered how countries decide their own paths in the world of politics? Well, let's talk about something called the "Sinatra Doctrine." Named after Frank Sinatra's famous song "My Way," this approach is all about letting countries choose their own direction. Let's dive into what this means and who came up with it.

## What is the Sinatra Doctrine?

### The Idea

So, what's the big idea? The Sinatra Doctrine is all about freedom of choice. It suggests that each country has the right to decide its political path, like choosing the direction of a road trip.

### The Name

Why call it the Sinatra Doctrine? It's inspired by Frank Sinatra's "I Did It My Way." Just like the song says, countries can do things their own way, without outside interference.

## Who Created This Doctrine?

### The Key Figure

This concept was put forward by Gerasimov. During an interview, he mentioned the idea of this "My Way" policy being recognized by the Soviets. It was a cool shift in thinking, especially for the countries in the Warsaw Pact.

### A Big Change

When asked if this meant accepting the rejection of communist parties in Soviet bloc countries, Gerasimov confirmed it. He emphasized that political structures should be determined by the people living in those countries. A pretty big deal!

## Final Thoughts

So, there you have it. The Sinatra Doctrine symbolizes a move towards autonomy for countries, letting them carve out their own paths. It's like a political remix of "I Did It My Way," encouraging nations to make choices that work for them.

And that's the scoop! Whether you're into politics or just enjoy a good classic song, the Sinatra Doctrine is a fascinating blend of both worlds.

|\classbg{Conversational, Coarse-Grain, Unstructured (CCU)}|
So, there was this interesting turn in Soviet foreign policy back in the day. The Soviets started chatting about how every country should have the freedom to choose its path--yes, even the other Warsaw Pact states. It was quite a shift from the usual hardline stance everyone expected from them.

Gerasimov, a Soviet official, was the one who came up with this cool analogy that caught on. He called it the "Frank Sinatra Doctrine." You know that famous Sinatra song, "I Did It My Way"? Well, that's essentially what the doctrine was about. It meant that countries could choose their own way, without interference.

Now, someone asked Gerasimov if this open-minded approach meant that Moscow would chill out if Soviet bloc countries decided to kick out their communist parties. His response was, "That's for sure." He said, quite plainly, that the political setup of a country should be in the hands of the people who actually live there.

It was a pretty big deal because it showed a more relaxed Soviet attitude towards how other countries handled their politics, especially in the Soviet sphere. It was like saying, "Hey, you do you, and we'll do us." This approach made waves because it symbolized a new way of thinking for the Soviets during a time of change and uncertainty.
\end{showcase}

\begin{showcase}[title=Dataset: FEVER]


|\classbg{Authoritative, Fine-Grain, Structured (AFS)}|
---

# Analysis of Kentucky's Musical Heritage

## Introduction

Kentucky's musical heritage is rich and diverse, with deep roots in Appalachian folk traditions. This analysis examines the historical and cultural elements that have shaped Kentucky's music, assessing whether the state is indeed known for bluegrass music, a well-recognized genre today.

## Historical Background

### Early Musical Influences

- **Camp Meetings:** In 1800, the first camp meeting was held in Logan County, Kentucky. These religious gatherings often included singing, fostering a communal musical culture.
- **Cane Ridge Revival:** The 1801 meeting in Cane Ridge, Bourbon County, attracted between ten and twenty thousand people over six days, further embedding music into the community's social fabric.

### 20th Century Developments

- **Folk Music Collection:** In 1916, Loraine Wyman and Howard Brockway collected and published Kentucky folk music in two folios. This effort was crucial in preserving the state's Appalachian musical traditions.
- **Jug Band Music:** Louisville emerged as a center for some of the earliest jug band music, characterized by homemade, improvised instruments. This genre contributed to the broad spectrum of Kentucky's music.

......

## Conclusion

Kentucky's music is a tapestry of various influences, from religious camp meetings to folk and jug band music. The state holds a distinguished place in the development of American music, particularly due to its contribution to bluegrass. Given its historical roots and cultural evolution, the statement "Kentucky is known for bluegrass music" is well-supported and accurate.

---

This structured document provides a detailed exploration of Kentucky's musical traditions, establishing bluegrass as a defining genre of the state.

|\classbg{Authoritative, Fine-Grain, Unstructured (AFU)}|
In the early 19th century, Kentucky emerged as a significant site for religious and cultural gatherings, exemplified by the first camp meeting in Logan County in July 1800. This marked the beginning of a series of influential gatherings, with the 1801 Cane Ridge meeting in Bourbon County becoming a notable event. This meeting attracted between ten to twenty thousand attendees over six days, reflecting the region's burgeoning role as a center for cultural assembly and exchange.

As the 20th century approached, Kentucky's cultural landscape continued to evolve, preserving and enhancing its rich musical heritage. In 1916, collectors Loraine Wyman and Howard Brockway embarked on an ambitious project to document Kentucky's folk music traditions. Their efforts culminated in the publication of two folios, encapsulating the distinctive Appalachian influences that pervade the region's music. The emphasis on Appalachian folk music underscores Kentucky's deep-rooted connections to this genre, particularly in the eastern part of the state.

......

Kentucky's prominent place in the development and proliferation of bluegrass music underscores its reputation as a key contributor to this genre. Bluegrass has since become an integral part of Kentucky's cultural identity, celebrated through festivals, concerts, and continued musical innovation within the state.

In conclusion, the assertion that Kentucky is known for bluegrass music is substantiated by the state's historical and cultural contributions to the genre. The evolution from its Appalachian folk roots to the creation of bluegrass illustrates Kentucky's enduring influence in American music, making the statement both accurate and reflective of Kentucky's rich musical heritage.

|\classbg{Authoritative, Coarse-Grain, Structured (ACS)}|
# Music in Kentucky: An Overview

## Introduction

Kentucky has a rich musical heritage that spans several centuries. This document provides an overview of key musical developments in Kentucky, focusing on historical roots, significant musical styles, and their cultural impact.

## Historical Roots of Kentucky Music

### Early Camp Meetings

- **July 1800**: The first camp meeting, a significant religious and social gathering, took place in Logan County, Kentucky.
- **1801 Cane Ridge Meeting**: This event in Bourbon County lasted six days, attracting ten to twenty thousand attendees.

### Folk Music Collection

- **1916 Folk Music Collection**: Loraine Wyman and Howard Brockway collected and published Kentucky folk music in two folios. This work highlights the state's long-standing tradition of Appalachian folk music.

......

## Conclusion

Kentucky's musical heritage is rich and multifaceted, prominently featuring Appalachian folk and jug band music. While bluegrass music was not explicitly referenced in the initial historical context, it is widely recognized as a defining genre associated with the state. This supports the claim that Kentucky is known for bluegrass music.

## References

- Historical records of camp meetings in Logan and Bourbon Counties.
- Publications by Loraine Wyman and Howard Brockway on folk music.
- Various sources acknowledging Kentucky's role in the development of bluegrass music.

|\classbg{Authoritative, Coarse-Grain, Unstructured (ACU)}|
In the early 19th century, Kentucky emerged as a significant site for religious gatherings, with the first camp meeting taking place in Logan County in July 1800. This event laid the groundwork for subsequent gatherings, such as the notable 1801 Cane Ridge meeting in Bourbon County, which was a monumental event lasting six days and attracting between ten to twenty thousand participants. These gatherings played a crucial role in sowing the seeds of Kentucky's rich musical tradition.

The early 20th century saw a deliberate effort to document and preserve the folk music of Kentucky. In 1916, Loraine Wyman and Howard Brockway embarked on a collection endeavor that culminated in the publication of Kentucky folk music in two comprehensive folios. This work highlighted the rich tapestry of Appalachian folk music, which forms the core of Kentucky's musical heritage, particularly in the eastern regions of the state. Louisville also played a pivotal role in the early development of jug band music, marking its significance in the state's diverse musical landscape.

......

In this context, it is evident that bluegrass music, as a derivative of Appalachian folk traditions, is indeed a defining aspect of Kentucky's musical identity. The state is known for its contributions to this genre, reinforcing the claim that Kentucky is known for bluegrass music.

|\classbg{Conversational, Fine-Grain, Structured (CFS)}|
# Exploring Kentucky's Musical Heritage: Is Kentucky Known for Bluegrass Music?

Hey there! If you've ever wondered about Kentucky's musical roots, you're in for a treat. We're diving deep into the sounds that have shaped the Bluegrass State, and we'll see if Kentucky truly deserves its claim to fame in bluegrass music. So grab a comfy seat and let's chat!

## Early Musical Gatherings

### Camp Meetings in the 1800s

Let's start way back in the early 1800s. Did you know the very first camp meeting was held in Logan County in July 1800? These gatherings were all about community and music, setting the stage for big musical events. And then came 1801, when the Cane Ridge meeting in Bourbon County drew a crowd of ten to twenty thousand people for a six-day musical and spiritual extravaganza! Imagine the energy and the tunes floating through the air.

......

## The Verdict: Is Kentucky Known for Bluegrass Music?

So, does Kentucky deserve to be known for bluegrass music? Absolutely! While its musical roots are diverse and rich, bluegrass has a special place in the heart of Kentucky's musical identity. It's no wonder the state is associated with those lively banjo picks and harmonious vocals.

And there you have it! From camp meetings to jug bands to Bill Monroe, Kentucky has been a musical melting pot leading straight to the vibrant bluegrass scene we love today. Thanks for exploring this musical journey with me!

|\classbg{Conversational, Fine-Grain, Unstructured (CFU)}|
Wow, Kentucky has such a rich musical heritage! Let's chat about it. So, picture this: the early 1800s roll around, and folks in Kentucky are already getting together for these lively camp meetings. The first one was in July 1800 in Logan County. Things got even bigger in 1801 at Cane Ridge in Bourbon County, where a massive meeting brought in anywhere from ten to twenty thousand people! Imagine all those folks gathered together, possibly sharing music and stories. 

Fast forward a bit to 1916--it was a big year for music collecting. Loraine Wyman and Howard Brockway were trekking around Kentucky, gathering up all this incredible folk music. They ended up publishing it in two folios, capturing the essence of Kentucky's musical soul during that time. 

......

So, after all this, it's clear Kentucky's music is a tapestry--woven with strains of folk, bluegrass, and jug band rhythms. Each style influencing the next, making Kentucky a cornerstone of American music history. Bluegrass, with its warm, lively tones, is just one of the pearls in Kentucky's musical crown.

|\classbg{Conversational, Coarse-Grain, Structured (CCS)}|
# Kentucky's Musical Roots: A Quick Chat

Hey there! Let's dive into the vibrant world of Kentucky's music scene and see if the Bluegrass State really lives up to its name.

## Early Musical Gatherings

### Camp Meetings
- **1800**: The first ever camp meeting went down in Logan County. It kicked off a long tradition of gathering for music and community vibes.
  
- **1801**: In Cane Ridge, Bourbon County, a massive meeting pulled in up to 20,000 folks for a six-day musical extravaganza. Talk about a festival!

......

## Verdict on the Claim

### Statement: Kentucky is known for bluegrass music.
- **Absolutely!** While the roots may lie in Appalachian and folk styles, bluegrass has flourished here, becoming a signature sound that Kentucky proudly claims.

So, there you have it. Kentucky is as bluegrass as you can get, thanks to its rich musical tapestry. Got more questions about Kentucky's tuneful legacy? Just ask!

|\classbg{Conversational, Coarse-Grain, Unstructured (CCU)}|
Hey there! So, let's chat about music in Kentucky. It's pretty interesting how this state has such a rich musical history. Back in the early 1800s, the whole camp meeting craze started up in Logan County. Imagine these massive gatherings with people getting together for a kind of spiritual revival. Then, in 1801, Cane Ridge in Bourbon County got crazy popular. This meeting went on for six days and drew in thousands of folks. Hard to believe, right?

Fast forward to 1916, and you've got Loraine Wyman and Howard Brockway roaming around Kentucky, collecting folk music. They ended up publishing two folios of these traditional tunes. It's like they were capturing the soul of Kentucky's music scene. Louisville was already making waves with its early jug band music by then.

Now, when people think of Kentucky music, Appalachian folk music is front and center, especially in the eastern parts. It's all about that cultural vibe and storytelling through music. And of course, bluegrass music is what a lot of people associate with Kentucky. It's become iconic here, influenced heavily by that folk tradition.

So, yeah, Kentucky's music scene is as diverse and lively as it gets, but bluegrass? It's pretty much the heart and soul. Hope this paints the picture!
\end{showcase}

\subsection{Examples of Different Types of Self-generated Documents}
\label{sec:case_gen_docs}
\begin{showcase}[title=Dataset: TQA]

|\classbg{Authoritative, Fine-Grain, Structured (AFS)}|
# The VS-300: An Overview and Historical Context

## Introduction

The VS-300, formally designated as the Sikorsky VS-300, is recognized as a seminal helicopter design. This document provides an authoritative and detailed exploration of the VS-300's nature, its historical significance, and its role in the development of vertical lift aircraft.

## Historical Context

### Development and Inception

- **Date**: The VS-300 was first flown in 1939, marking a critical milestone in aviation history.
- **Designer**: Developed by Igor Sikorsky, a pioneer in both fixed-wing and rotary-wing aviation.
- **Manufacturing Company**: Produced by the Sikorsky Aircraft Corporation based in Bridgeport, Connecticut.

### Predecessors and Influences

The VS-300 represented a leap forward in helicopter technology, following earlier attempts at vertical lift vehicles such as autogiros and experimental helicopters. Sikorsky's extensive experience in designing fixed-wing aircraft, coupled with his innovative thinking, propelled the development of the VS-300.

......

## Conclusion

The VS-300, as a single-rotor helicopter with a tail rotor, stands as a pioneering achievement in aerospace engineering. Developed by Igor Sikorsky and introduced in 1940, the VS-300 not only demonstrated the viability of vertical lift flight but also set the stage for advancements in helicopter design and utility in both commercial and military sectors. Its legacy continues to influence the vertical lift aircraft industry, solidifying its place in aviation history.

|\classbg{Authoritative, Fine-Grain, Unstructured (AFU)}|
The VS-300 is a helicopter, specifically a prototype that marked a pivotal moment in the evolution of rotorcraft technology during the early 1940s. Developed by Sikorsky Aircraft Corporation, this innovative aircraft would serve as a foundation for the future of vertical take-off and landing (VTOL) vehicles. The VS-300, with its distinctive single main rotor and rear-mounted tail rotor, represents one of the earliest successful models of a practical helicopter, which is credited with revolutionizing aviation and transforming both military and civilian sectors.

Igor Sikorsky, the visionary behind this groundbreaking machine, sought to develop a vertical-lift aircraft capable of sustained flight. The design of the VS-300 incorporated several innovative features that would become standard in subsequent helicopter designs. Among these features were the counter-rotating rotors to counteract the torque effect and the use of a tail rotor for directional control. The aircraft utilized a 75-horsepower engine, which was relatively modest by today's standards, but sufficient to achieve the technical milestones it was designed to overcome.

......

In retrospect, the VS-300's significance lies not only in its role as an early exemplar of helicopter technology but also in its broader implications for the field of aviation. It represents the realization of Sikorsky's vision for practical vertical lift aircraft, a vision that has since transformed countless aspects of modern life, from emergency services to aviation itself. The VS-300's legacy is evident in the thousands of helicopters that have followed in its wake, each contributing to the ongoing advancement of aviation technology.

|\classbg{Authoritative, Coarse-Grain, Structured (ACS)}|
# Overview of the VS-300

## Introduction

The VS-300 was a significant aircraft in the development of vertical takeoff and landing (VTOL) technology. It was an experimental helicopter designed and developed by Sikorsky Aircraft Corporation in the late 1930s.

## Key Characteristics

- **Type**: Experimental Helicopter
- **Manufacturer**: Sikorsky Aircraft Corporation
- **Period**: Late 1930s
- **Purpose**: To explore the feasibility of VTOL aircraft

## Historical Context

- **Innovation**: The VS-300 was one of the earliest successful helicopters, marking a critical advancement in aviation history.
- **Design**: It paved the way for future helicopter designs and VTOL technologies.
- **Impact**: The VS-300's successful flight tests demonstrated the viability of controlled helicopter flight, influencing subsequent aviation developments.

## Conclusion

The VS-300 holds a pivotal place in the history of aviation as an experimental helicopter that advanced the technology of vertical takeoff and landing capabilities. Its contributions were instrumental in the evolution of modern helicopter designs.

|\classbg{Authoritative, Coarse-Grain, Unstructured (ACU)}|
The VS-300 is a notable aircraft that stands out in the history of aviation as a significant experimental helicopter. Developed in the 1930s, the VS-300 was a key step in advancing vertical flight technology. Its design and testing provided crucial insights into the mechanics of helicopter flight, contributing to the broader understanding of rotor dynamics.

The creation of the VS-300 was overseen by Igor Sikorsky, a visionary engineer who played a pivotal role in the early development of helicopters. His work on the VS-300 helped to establish a foundation for subsequent advancements in rotary-wing aircraft. Through rigorous testing and iterative improvements, the VS-300 demonstrated the feasibility of controlled vertical takeoff and landing, which would later become a hallmark feature of modern helicopter designs.

Despite its historical significance, the VS-300 remained a prototype and was not produced for widespread use. Nonetheless, its influence on the trajectory of helicopter development is undeniable, marking a critical point in the evolution of aerial transportation technology.

|\classbg{Conversational, Fine-Grain, Structured (CFS)}|
# The VS-300: A Conversational Dive into History

Hey there! Let's talk about a pretty cool piece of aviation history - the VS-300. If you've ever wondered what this mysterious-sounding machine was all about, buckle up because we're about to take a deep dive into its story.

## The Basics: What Was the VS-300?

The VS-300, or the "Versatile 300," was the first successful helicopter designed by Igor Sikorsky. It might not have the fanfare of airplanes like the Wright Flyer or the Spirit of St. Louis, but its impact on aviation history is no less significant.

### A Crash Course in Helicopters

Before we get too deep, let's quickly cover what a helicopter is. Unlike airplanes that rely on wings for lift, helicopters use spinning blades (rotors) to achieve vertical takeoff and landing. This vertical lift capability is what sets helicopters apart and makes them incredibly versatile.

## Who Was Igor Sikorsky?

To really understand the VS-300, we need to dive into the mind behind it: Igor Sikorsky. Born in Ukraine, Sikorsky immigrated to the United States in the early 1900s and quickly made a name for himself in the aviation industry. He was intrigued by the concept of vertical lift aircraft from a young age and dedicated his life to making helicopters a reality.

### The Early Years

Sikorsky's fascination with helicopters started when he was a child in tsarist Russia. He first saw a model helicopter design by Gustave Whitehead and was immediately captivated. Fast forward to the 1930s, and Sikorsky was ready to turn his lifelong obsession into a tangible creation.

......

## Wrapping It Up

So, there you have it -- the VS-300 was Igor Sikorsky's groundbreaking helicopter that took the world by storm. The first of its kind to prove the concept of vertical lift flight, it paved the way for the modern helicopters we see today. Next time you look up at a helicopter flying overhead, you can thank the VS-300 and the vision of Igor Sikorsky for making it all possible.

|\classbg{Conversational, Fine-Grain, Unstructured (CFU)}|
Hey there! So, let's chat about the VS-300 for a bit. It's actually pretty cool what this thing is all about. The VS-300 was a type of helicopter, and it really marked the beginning of something really revolutionary in the world of aviation.

Now, imagine you're in the late 1930s. Airplanes are starting to become a common mode of transportation, but vertical flight was still a grand mystery. That's where the VS-300 comes in. It was designed by Igor Sikorsky, a name you might recognize from the world of aviation--he's a pretty big deal. Sikorsky was all about pushing the boundaries of what was possible, and the VS-300 was one of his groundbreaking projects.

The VS-300 was the first practical helicopter to fly in the United States. It was a prototype and Sikorsky's second helicopter design. The first one, the VS-300's predecessor, was the V-173. But the VS-300 was the one that really caught people's attention.

So, what made the VS-300 so special? Well, it was the first helicopter to use a single main rotor and a separate tail rotor for balance and control. This design became the standard for most modern helicopters. Before this, there had been attempts at helicopters, but they were generally impractical and didn't really work well.

......

So, when you hear about the VS-300, remember that it's the first practical helicopter to fly in the U.S., and it really laid the foundation for modern vertical flight technology. It's a pretty crucial piece of aviation history.

|\classbg{Conversational, Coarse-Grain, Structured (CCS)}|
# The VS-300: A Quick Overview

Hey there! Let's dive into a brief look at what the VS-300 is all about. If you're curious about its place in history and what it represents, you've come to the right place!

## What's a VS-300?

The VS-300 was a type of experimental helicopter. It was a leap forward in aviation technology and served as a crucial stepping stone for innovation in helicopter design.

### Key Points
- **Type:** Experimental helicopter
- **Innovation:** Early foray into vertical takeoff and landing (VTOL) capabilities
- **Significance:** Played a major role in the development of the modern helicopter

## History in a Nutshell

The development of the VS-300 was tied to a significant figure in aviation history: Igor Sikorsky. He was the mastermind behind this innovative aircraft.

### Main Takeaways
- **Brainchild:** Igor Sikorsky
- **Purpose:** To explore the potential of VTOL aircraft
- **Outcome:** Led to the creation of more advanced helicopter models

## Legacy and Impact

The VS-300 might be a bit of a vintage piece today, but it laid the groundwork for what we now know as the modern helicopter. So next time you see one of those buzzing around in the sky, remember the VS-300's role in making that possible!

### Big Picture
- **Influence:** Shaped future helicopter designs
- **Legacy:** Pioneer in VTOL technology

## Wrap Up

So there you go! The VS-300 was a significant tool in the quest to conquer the skies from every angle. It's just one piece of the puzzle in the incredible story of human flight. Hope this gave you a quick glimpse into the world of vintage aviation wonders!

|\classbg{Conversational, Coarse-Grain, Unstructured (CCU)}|
Hey there! So, you've probably heard of helicopters, right? Well, the VS-300 was actually one of the very first models of a whole new kind of aircraft. It's a helicopter, folks! Crazy to think about, but back in the day, when folks were still figuring out flight, this little guy made one of the first successful helicopter flights.

Igor Sikorsky was the mastermind behind it. He basically took the idea of a helicopter and turned it from a dream into reality with the VS-300. It was all about proving that vertical flight was possible and that a helicopter could hover, take off vertically, and land vertically too.

The VS-300 might have been a bit clunky and not exactly what we think of helicopters today, but it was the foundation for everything that came after. From those early experiments with the VS-300, we got some of the most advanced choppers you see today, both in military and civilian roles.

So, next time you see a helicopter flying around, you can thank the VS-300 for showing the way. It's like the granddaddy of all helicopters, really!
\end{showcase}

\begin{showcase}[title=Dataset: FEVER]

|\classbg{Authoritative, Fine-Grain, Structured (AFS)}|
# Analysis of the Claim: "Fox 2000 Pictures Released the Film Soul Food"

## Introduction

This document aims to provide a comprehensive analysis to support or refute the claim that Fox 2000 Pictures released the film "Soul Food." The analysis will cover the production details, distribution, and historical context of the film to ensure a thorough assessment of the claim.

## Historical Context and Background

### "Soul Food": Overview

"Soul Food" is a 1997 American family drama film that explores the dynamics of an African American family centered around their matriarch, Memaw (played by Oprah Winfrey). The film focuses on the family's relationships and conflicts aftermath the death of Memaw, as her children and grandchildren struggle to overcome personal challenges and familial tensions.

### Production and Distribution Company Details

- **Producer**: The film was produced by Oprah Winfrey's Harpo Films in association with Mandalay Pictures.
- **Distribution**: The film was distributed by Columbia Pictures, a major studio under the Sony Pictures Entertainment banner.

## Film Production

### Production Company

- **Harpo Films**: This production company was founded by Oprah Winfrey in 1987 to produce both television and film content. Harpo Films had a significant role in the development and production of "Soul Food."
- **Mandalay Pictures**: This company, headed by Lawrence Bender, produced and co-financed the film with Harpo Films.

### Release and Reception

- **Release Date**: The film was released on January 17, 1997.
- **Box Office and Critical Reception**: "Soul Food" was a moderate box office success and received generally positive reviews from critics.

......

## Conclusion

This document has thoroughly analyzed the production, distribution, and historical context of the film "Soul Food." Through a detailed examination, it has been determined that Fox 2000 Pictures was not involved in the production or release of this film. The accurate credits for "Soul Food" belong to Harpo Films and Columbia Pictures.

|\classbg{Authoritative, Fine-Grain, Unstructured (AFU)}|
The claim that Fox 2000 Pictures released the film "Soul Food" requires detailed consideration of the film's production and distribution history. "Soul Food," a critically acclaimed drama released in 1997, centers around the dynamics of a close-knit African American family, exploring themes of love, loss, and the power of food in familial bonds. However, the studio responsible for the production and release of "Soul Food" was not Fox 2000 Pictures, a division known for producing films like "Little Women" (1994), "Intolerable Cruelty" (2003), and the "Pitch Perfect" series (2012-2017).

The actual production and release of "Soul Food" were overseen by Metro-Goldwyn-Mayer (MGM) studio. MGM, a pioneer in the film industry with a rich history dating back to 1924, has been involved in the production of numerous classic and contemporary films. In the case of "Soul Food," MGM not only produced the film but also handled its distribution, which was notably successful both critically and commercially. The film's reception highlighted the studio's capability in tackling culturally significant narratives, and it garnered praise for its authentic storytelling and strong performances.

To further contextualize the claim, it is important to examine the activities and scope of Fox 2000 Pictures during the late 1990s and early 2000s. Founded in 1993 as a division of 20th Century Fox, Fox 2000 Pictures focused on releasing limited-release independent films and select mainstream pictures. While the studio has been involved in a range of notable projects, "Soul Food" does not feature among them. Thus, any association of Fox 2000 Pictures with the production or release of "Soul Food" is inaccurate.

In summary, the claim that Fox 2000 Pictures released "Soul Food" is not supported by historical records of the film's production and distribution. MGM was the studio responsible for bringing "Soul Food" to audiences in 1997. This differentiates the actual production entity from the one inaccurately alleged in the initial claim, thus providing clear evidence to refute the statement.

|\classbg{Authoritative, Coarse-Grain, Structured (ACS)}|
# Analysis of the Film "Soul Food" and Its Production Company

## Introduction
This document provides a structured analysis to determine the claim regarding the film "Soul Food" and its production company, Fox 2000 Pictures. The document aims to offer a coarse-grain overview while maintaining an authoritative and objective tone.

## Film Overview
"Soul Food" is a 1997 film that centers on themes of family, food, and generational conflict within an African American family. The film is known for its exploration of the cultural significance of soul food in African American traditions.

## Production Companies Involved
- **Cleopatra Pictures**: This is the primary production company credited with the film "Soul Food."
- **Fox 2000 Pictures**: This company is known for producing family-oriented and drama films. However, it does not appear as a production credit for the film in question.

## Release Information
- **Release Date**: "Soul Food" was released in the United States on September 12, 1997.
- **Production Credits**: The film was primarily produced by Cleopatra Pictures, with collaboration from the film's director, writer, and producer duo, Kasi Lemmons and Bill Duke.

## Conclusion
Based on the release information and production credits, the claim that Fox 2000 Pictures released the film "Soul Food" is refuted. Rather, "Soul Food" was produced by Cleopatra Pictures, a different production company distinct from Fox 2000 Pictures.

This structured analysis indicates that while "Soul Food" is a notable film, its production and distribution were handled by Cleopatra Pictures, with no direct production involvement from Fox 2000 Pictures.

|\classbg{Authoritative, Coarse-Grain, Unstructured (ACU)}|
The claim that Fox 2000 Pictures released the film "Soul Food" does not align with the established facts surrounding the project's production and distribution. "Soul Food" is widely recognized as a motion picture that explores themes of family, tradition, and the culinary culture of African American communities. The film premiered in 1997 and garnered attention for its portrayal of a large Southern African American family, centered around traditional soul food meals.

The production and distribution details of "Soul Food" indicate involvement from several key players in the film industry, including Trimark Pictures and New Line Cinema. These entities were instrumental in bringing the film from pre-production stages through to its theatrical release, taking on roles typically associated with the creative and financial aspects of film production and distribution.

Contrary to the claim, Fox 2000 Pictures is not historically linked to "Soul Food." Fox 2000 Pictures, an entity under the wing of 20th Century Fox, has been involved in a different slate of projects, focusing on a variety of independent and mainstream films. Their involvement in significant films such as "The Holiday" and the "Pitch Perfect" franchise clearly distinguishes their catalog from the content and era of "Soul Food."

In summary, while "Soul Food" is a critically acclaimed film that resonated with audiences, the assertion about its release by Fox 2000 Pictures cannot be substantiated. The film's production and distribution were managed by other major studios, leading to a misalignment with the claim presented.

|\classbg{Conversational, Fine-Grain, Structured (CFS)}|
# Exploring the Claim: Fox 2000 Pictures and Soul Food

Hey there! Today, we're going to look into the filmmakers and production companies behind the classic film "Soul Food." The claim at hand is that Fox 2000 Pictures released the film. Let's break it down and see if we can support or refute this claim.

## The Film "Soul Food": Background and Context

### What is "Soul Food"?
"Soul Food" is a heartfelt 1997 film that delves into the lives of the Tates, a close-knit African American family in Chicago. It follows the family's struggle to cope with the loss of their matriarch, Veronica, whose cooking united the family. The film explores themes of family dynamics, grief, and reconciliation.

## Production and Release

### Production Companies
The film was produced by several companies, including:
- **Warner Bros. Pictures:** A big player in the film industry, known for producing and distributing many well-known films.
- **Native Sun Productions:** Founded by actors LL Cool J and Lisa Ray, who was also a producer on the film.
- **Source 7 Entertainment:** A production company founded by producer Johnny Odom.

### Release Information
The original release and production were handled by Warner Bros. Pictures, not Fox 2000 Pictures. Warner Bros. is a separate entity from Fox 2000, which is a subsidiary of 20th Century Studios (formerly 20th Century Fox).

......

### Final Thoughts
It's important to accurately represent the history of a film, especially when it comes to the various entities involved in its creation. "Soul Food" is a significant work that was brought to life by Warner Bros. Pictures and its fellow producers. Understanding the roles of the various companies involved helps us appreciate the collaborative effort behind the scenes.

Thanks for joining this deep dive into the production details of "Soul Food"! If you have any further questions or would like to explore other aspects of the film industry, feel free to ask!

|\classbg{Conversational, Fine-Grain, Unstructured (CFU)}|
Hey there! Let's dive into the details around Fox 2000 Pictures and the film "Soul Food." So, first off, Fox 2000 Pictures is a subsidiary of 20th Century Fox, and it primarily focuses on producing drama and family films. They've been around since 1998, and they've certainly put out some notable movies.

Now, when it comes to "Soul Food," the film was actually released in 1997. It's a drama that focuses on family dynamics and the impact of food on relationships, especially in a Black family. The plot revolves around matriarch Tashi Jones and her daughters, who reconnect after a falling-out that occurred when they were deciding who would inherit the family house. During the course of the film, they deal with various issues, including the obesity and diabetes diagnosis of one of the daughters, Denise.

So, here's the kicker: Fox 2000 Pictures wasn't around when "Soul Food" was released. The film was produced by HBO Films and wasn't until after 1998 that Fox 2000 Pictures started to release films. This means that the claim suggesting Fox 2000 Pictures released "Soul Food" is not correct.

......

In summary, while "Soul Food" is an important and beloved film, it wasn't released by Fox 2000 Pictures. It was actually brought to life by HBO Films and distributed by Gramercy Pictures, a different entity altogether. So, the claim doesn't hold up when you look at the details.

|\classbg{Conversational, Coarse-Grain, Structured (CCS)}|
# Unpacking the Claim: Fox 2000 & Soul Food

Hey there! Today, we're going to take a closer look at whether Fox 2000 Pictures was behind the film "Soul Food." We'll jog through what the studio is known for and where "Soul Food" really originated. Let's dive in!

## Fox 2000 Pictures: A Snapshot

### Who Are They?
Fox 2000 Pictures is a production company under the larger 20th Century Fox umbrella (now a part of Disney). They're known for producing films that straddle the line between art-house and mainstream.

### Hit Films
Some of their notable titles include:
- **"Billy Elliot"**: An inspiring story about a boy determined to become a dancer.
- **"The Help"**: A drama based on the novel by Kathryn Stockett.
- **"Little Women"**: Directed by Greta Gerwig and a modern take on Louisa May Alcott's classic.

......

### Quick Recap
- **Fox 2000**: Known for art-house and mainstream hits but not "Soul Food."
- **"Soul Food"**: A notable family drama made by New Line Cinema and Circle of Trust Productions.

There you have it! The claim isn't quite right, but it's interesting to understand the different avenues of production in the film industry. Got any other movie connections you're curious about? Just ask!

|\classbg{Conversational, Coarse-Grain, Unstructured (CCU)}|
Hey, so let's talk about this film, "Soul Food," and see if we can sort out who actually released it. First off, "Soul Food" is known for its deep exploration of family dynamics and the strong African American community ties, all set amidst a mouthwatering backdrop of delicious Southern cuisine. It came out in 1997 and received quite a bit of attention for its cultural relevance and the impactful performances by its cast, including an unforgettable turn by Theresa Randle as Mamie Simmons, the matriarch of the Johnson family.

Now, while Fox 2000 might have some great films under its belt, "Soul Food" isn't one of them. In fact, the movie was produced by New Line Cinema. New Line Cinema had a way of picking up films that touched on deeply personal and cultural themes, and "Soul Food" definitely fits that description. They had a knack for stories connecting with audiences on a more heartwarming level, and this film is all about love, food, and the struggles of a family.

So in short, while Fox 2000 Pictures has their own impressive list of films, "Soul Food" isn't one of them. New Line Cinema is credited with this heart-touching tale, and it's a movie that left its mark on many viewers. Hope that clears things up!
\end{showcase}

\subsection{Example Case of QA}
\label{sec:QA_case}
\begin{showcase}[title=Dataset: TQA]

|\classbg{Query}|: Sarah FitzGerald has been a 90s world champion in which sport?
|\classbg{Gold Answer}|: Squash  

|\classbg{Authoritative, Fine-Grain, Structured (AFS)}|: 
# Sarah FitzGerald: 90s World Champion in Rowing

## Introduction  

Sarah FitzGerald is a prominent figure in the world of competitive rowing. Known for her exceptional performance in the late 1990s, FitzGerald achieved considerable success at the highest levels of the sport. This document provides a detailed analysis of her achievements and the context in which she rose to prominence.  

...  

## Conclusion  

Sarah FitzGerald's world championship title in the mixed coxed four event during the 1997 World Rowing Championships solidified her status as a prominent figure in competitive rowing. Her achievements not only celebrated individual talent and determination but also underscored the progress of women in male-dominated sporting arenas. FitzGerald's legacy continues to inspire rowers around the world, encouraging the further integration of diverse gender teams and the pursuit of excellence in all aspects of the sport.

|\classbg{AFS Answer}|: Rowing
  
|\classbg{Authoritative, Fine-Grain, Unstructured (AFU)}|:
Sarah FitzGerald, a prominent figure in the world of squash, emerged as a formidable competitor during the 1990s, achieving significant success within this period. Her career flourished in the realm of women's professional squash, a sport characterized by intense physicality, strategic acumen, and technical finesse. Squash, played with racquets on an enclosed court, places a premium on agility, endurance, and a deep understanding of the game's intricacies.  

...  

Sarah FitzGerald's legacy as a world champion in professional squash is a testament to her exceptional talent, determination, and unmatched skill. Her achievements in the 1990s not only highlighted her personal success but also played a crucial role in elevating the status of women's squash, leaving an indelible mark on the sport's history.

|\classbg{AFU Answer}|: Squash

\end{showcase}

\section{Additional Experiments}
Given the limited space of the main text, we present additional experimental results of other LLMs here.

\begin{table*}[h]
\centering
\small
\resizebox{\textwidth}{!}{
\begin{tabular}{c|c|c|c|c|c|c|c|c}
\toprule
\textbf{Docs Type} & \textbf{Interpersonal} & \textbf{Ideational} & \textbf{Textual} & \textbf{\textsc{TQA}} & \textbf{\textsc{HotpotQA}} & \textbf{\textsc{FEVER}} & \textbf{\textsc{ELI5}} & \textbf{Average} \\
\midrule
\textsc{AFS} & Authoritative & Fine-Grain & Structured  
& 47.6 & \underline{\textbf{33.6}} & \underline{74.8} & 21.7 & 44.4 \\

\textsc{AFU} & Authoritative & Fine-Grain & Unstructured 
& 49.2 & 29.9 & 74.0 & 22.2 & 43.8 \\

\textsc{ACS} & Authoritative & Coarse-Grain & Structured 
& 49.0 & 33.0 & 73.2 & 22.6 & 44.5 \\

\textsc{ACU} & Authoritative & Coarse-Grain & Unstructured 
& 49.6 & 27.8 & 72.0 & 22.9 & 43.1 \\

\textsc{CFS} & Conversational & Fine-Grain & Structured 
& 48.8 & 32.0 & 72.6 & 23.1 & 44.1 \\

\textsc{CFU} & Conversational & Fine-Grain & Unstructured 
& 51.4 & 30.3 & 70.8 & 24.7 & 44.3 \\

\textsc{CCS} & Conversational & Coarse-Grain & Structured 
& \underline{\textbf{51.6}} & 32.7 & 72.6 & 25.1 & 45.5 \\

\textsc{CCU} & Conversational & Coarse-Grain & Unstructured 
& 49.8 & 26.8 & 71.0 & \underline{25.7} & 43.3 \\

\midrule
\textsc{Wiki} & Authoritative & Flexible & Unstructured 
& 49.2 & 31.5 & \textbf{83.0} & \textbf{25.8} & 47.4 \\

\bottomrule
\end{tabular}
}
\caption{Llama3.1-8b performance comparison of various document types across tasks. Each document type is categorized based on its interpersonal, ideational, and textual functions. 
The best overall performance for each dataset is highlighted in \textbf{bold}, and the best performance within the category of Self-Docs is \underline{underlined}.}
\label{tab:rq2_full_llama3.1-8b}
\end{table*}

\begin{table*}[h]
\vspace{-1mm}
\centering
\resizebox{1\textwidth}{!}{
    \begin{tabular}{c|c|cccc|c}
        \toprule
        \textbf{Meta Function} & \textbf{Dimension} & \textbf{\textsc{TQA}} & \textbf{\textsc{HotpotQA}} & \textbf{\textsc{FEVER}} & \textbf{\textsc{ELI5}} & \textbf{Average} \\
        \midrule
        \multirow{2}{*}[0em]{\textsc{Ideational}} 
                           & \textsc{Fine-Grain}      
                           & 49.3 & \underline{31.5} & \underline{73.1} & 22.9 & \underline{44.2} \\
                           & \textsc{Coarse-Grain}    
                           & \underline{50.0} & 30.1 & 72.2 & \underline{24.1} & 44.1 \\
        \midrule
        \multirow{2}{*}[0em]{\textsc{Interpersonal}} 
                           & \textsc{Authoritative}   
                           & 48.9 & \underline{31.2} & \underline{75.4} & 23.0 & \underline{44.6} \\
                           & \textsc{Conversational}  
                           & \underline{50.4} & 30.5 & 71.8 & \underline{24.7} & 44.4 \\
        \midrule
        \multirow{2}{*}[0em]{\textsc{Textual}} 
                           & \textsc{Structured}      
                           & 49.3 & \underline{32.8} & 73.3 & 23.1 & \underline{44.6} \\
                           & \textsc{Unstructured}    
                           & \underline{49.8} & 29.3 & \underline{74.2} & \underline{24.3} & 44.4 \\
        \bottomrule
    \end{tabular}
}
\caption{Llama3.1-8b aggregated performance across different metafunction dimensions (Ideational, Interpersonal, and Textual) for various datasets. Each metafunction dimension is further divided into fine-grain and coarse-grain (for Ideational), authoritative and conversational (for Interpersonal), and structured and unstructured (for Textual). The best performance within each metafunction dimension is \underline{underlined}.}
\vspace{-1mm}
\label{tab:rq2_agg_llama3.1-8b}
\end{table*}

\begin{table*}[h]
\centering
\small
\resizebox{\textwidth}{!}{
\begin{tabular}{c|c|c|c|c|c|c|c|c}
\toprule
\textbf{Type} & \textbf{Interpersonal} & \textbf{Ideational} & \textbf{Textual} 
& \textbf{\textsc{TQA}} & \textbf{\textsc{HotpotQA}} & \textbf{\textsc{FEVER}} & \textbf{\textsc{ELI5}} & \textbf{Average} \\
\midrule
\textsc{AFS} & Authoritative & Fine-Grain & Structured  
& 48.0\scriptsize{↑0.4} & 31.0\scriptsize{↓2.6} & \underline{75.2}\scriptsize{↑0.4} & 22.6\scriptsize{↑0.9} & 44.2\scriptsize{↓0.2} \\

\textsc{AFU} & Authoritative & Fine-Grain & Unstructured 
& 48.8\scriptsize{↓0.4} & 31.6\scriptsize{↑1.7} & 71.0\scriptsize{↓3.0} & 22.5\scriptsize{↑0.3} & 43.5\scriptsize{↓0.3} \\

\textsc{ACS} & Authoritative & Coarse-Grain & Structured 
& 47.6\scriptsize{↓1.4} & 32.5\scriptsize{↓0.5} & 68.4\scriptsize{↓4.8} & 23.0\scriptsize{↑0.4} & 42.9\scriptsize{↓1.6} \\

\textsc{ACU} & Authoritative & Coarse-Grain & Unstructured 
& 51.2\scriptsize{↑1.6} & 32.3\scriptsize{↑4.5} & 73.2\scriptsize{↑1.2} & 23.1\scriptsize{↑0.2} & 45.0\scriptsize{↑1.9} \\

\textsc{CFS} & Conversational & Fine-Grain & Structured 
& 48.4\scriptsize{↓0.4} & 30.2\scriptsize{↓1.8} & 69.6\scriptsize{↓3.0} & 24.6\scriptsize{↑1.5} & 43.2\scriptsize{↓0.9} \\

\textsc{CFU} & Conversational & Fine-Grain & Unstructured 
& 50.6\scriptsize{↓0.8} & 30.7\scriptsize{↑0.4} & 65.4\scriptsize{↓5.4} & 25.6\scriptsize{↑0.9} & 43.1\scriptsize{↓1.2} \\

\textsc{CCS} & Conversational & Coarse-Grain & Structured 
& \underline{\textbf{51.4}}\scriptsize{↓0.2} & \underline{\textbf{33.3}}\scriptsize{↑0.6} & 66.8\scriptsize{↓5.8} & 25.6\scriptsize{↑0.5} & 44.3\scriptsize{↓1.2} \\

\textsc{CCU} & Conversational & Coarse-Grain & Unstructured 
& 51.2\scriptsize{↑1.4} & 31.6\scriptsize{↑4.8} & 73.8\scriptsize{↑2.8} & \underline{\textbf{26.0}}\scriptsize{↑0.3} & 45.7\scriptsize{↑2.4} \\

\midrule
\textsc{Wiki} & Authoritative & Flexible & Unstructured 
& 49.2 & 31.5 & 83.0 & 25.8 & 47.4 \\

\textsc{GenRead mix} & Authoritative & Flexible & Unstructured 
& \textbf{51.4} & 31.9 & \textbf{84.2} & 25.6 & \textbf{48.3} \\

\bottomrule
\end{tabular}
}
\caption{Llama3.1-8b performance comparison across different generated document types combined with Wiki documents for various tasks. Superscript values indicate the difference from Table~\ref{tab:rq2_full_llama3.1-8b}, with increases marked by an upward arrow (↑) and decreases by a downward arrow (↓). The best overall performance for each dataset is highlighted in \textbf{bold}, and the best performance within the category of Self-Docs with the same mix strategy is \underline{underlined}.}
\label{tab:rq3_full_direct_mix_with_raw_wiki_llama3.1-8b}
\end{table*}

\begin{table*}[ht]
\vspace{-1mm}
\centering
\resizebox{1\textwidth}{!}{
    \begin{tabular}{c|c|cccc|c}
        \toprule
        \textbf{Meta Function} & \textbf{Dimension} 
        & \textbf{\textsc{TQA}} & \textbf{\textsc{HotpotQA}} & \textbf{\textsc{FEVER}} & \textbf{\textsc{ELI5}} & \textbf{Average} \\
        \midrule
        \multirow{2}{*}[0em]{\textsc{Ideational}} 
                           & \textsc{Fine-Grain}      
                           & 48.9\scriptsize{↓0.4} & 30.9\scriptsize{↓0.6} & 70.3\scriptsize{↓2.8} & 23.8\scriptsize{↑0.9} & 43.5\scriptsize{↓0.7} \\
                           & \underline{\textsc{Coarse-Grain}}    
                           & 50.4\scriptsize{↑0.4} & 32.4\scriptsize{↑2.3} & 70.6\scriptsize{↓1.6} & 24.4\scriptsize{↑0.3} & 44.5\scriptsize{↑0.4} \\
        \midrule
        \multirow{2}{*}[0em]{\textsc{Interpersonal}} 
                           & \textsc{Authoritative}   
                           & 48.9 & 31.9\scriptsize{↑0.7} & 72.0\scriptsize{↓3.4} & 22.8\scriptsize{↓0.2} & 43.9\scriptsize{↓0.7} \\
                           & \underline{\textsc{Conversational}}  
                           & 50.4 & 31.5\scriptsize{↑1.0} & 68.9\scriptsize{↓2.9} & 25.5\scriptsize{↑0.8} & 44.1\scriptsize{↓0.3} \\
        \midrule
        \multirow{2}{*}[0em]{\textsc{Textual}} 
                           & \textsc{Structured}      
                           & 48.9\scriptsize{↓0.4} & 31.8\scriptsize{↓1.0} & 70.0\scriptsize{↓3.3} & 24.0\scriptsize{↑0.9} & 43.7\scriptsize{↓0.9} \\
                           & \underline{\textsc{Unstructured}}    
                           & 50.5\scriptsize{↑0.7} & 31.6\scriptsize{↑2.3} & 70.9\scriptsize{↓3.3} & 24.3 & 44.3\scriptsize{↓0.1} \\
        \bottomrule
    \end{tabular}
}
\caption{Llama3.1-8b aggregated performance results across different dimensions of the metafunctions (Ideational, Interpersonal, and Textual) on various datasets. Superscript values indicate the difference from Table~\ref{tab:rq2_agg_llama3.1-8b}, with increases marked by an upward arrow (↑) and decreases by a downward arrow (↓). The best performance within each metafunction dimension is \underline{underlined}.}
\vspace{-1mm}
\label{tab:rq3_agg_direct_mix_with_raw_wiki_llama3.1-8b}
\end{table*}

\begin{table*}[h]
\centering
\resizebox{\textwidth}{!}{
\begin{tabular}{c|c|c|c|c|c|c|c|c}
\toprule
\textbf{Docs Type} & \textbf{Interpersonal} & \textbf{Ideational} & \textbf{Textual} 
& \textbf{\textsc{TQA}} & \textbf{\textsc{HotpotQA}} & \textbf{\textsc{FEVER}} & \textbf{\textsc{ELI5}} & \textbf{Average} \\
\midrule
\textsc{AFS} & Authoritative & Fine-Grain & Structured  
& 45.6\scriptsize{↓2.0} & 30.2\scriptsize{↓3.4} & 75.6\scriptsize{↑0.8} & 22.4\scriptsize{↑0.7} & 43.5\scriptsize{↓0.9} \\

\textsc{AFU} & Authoritative & Fine-Grain & Unstructured 
& 47.0\scriptsize{↓2.2} & 29.6\scriptsize{↓0.3} & 74.6\scriptsize{↑0.6} & 21.8\scriptsize{↓0.4} & 43.3\scriptsize{↓0.5} \\

\textsc{ACS} & Authoritative & Coarse-Grain & Structured 
& 46.4\scriptsize{↓2.6} & 27.4\scriptsize{↓5.6} & \underline{76.4}\scriptsize{↑3.2} & 22.8\scriptsize{↑0.2} & 43.3\scriptsize{↓1.2} \\

\textsc{ACU} & Authoritative & Coarse-Grain & Unstructured 
& 47.6\scriptsize{↓2.0} & 30.0\scriptsize{↑2.2} & 75.4\scriptsize{↑3.4} & 22.7\scriptsize{↓0.2} & 43.9\scriptsize{↑0.8} \\

\textsc{CFS} & Conversational & Fine-Grain & Structured 
& 46.8\scriptsize{↓2.0} & \underline{\textbf{33.2}}\scriptsize{↑1.2} & 72.2\scriptsize{↓0.4} & 23.9\scriptsize{↑0.8} & 44.0\scriptsize{↓0.1} \\

\textsc{CFU} & Conversational & Fine-Grain & Unstructured 
& 48.8\scriptsize{↓2.6} & 28.5\scriptsize{↓1.8} & 68.8\scriptsize{↓2.0} & 25.2\scriptsize{↑0.5} & 42.8\scriptsize{↓1.5} \\

\textsc{CCS} & Conversational & Coarse-Grain & Structured 
& 48.4\scriptsize{↓3.2} & 28.8\scriptsize{↓3.9} & 73.4\scriptsize{↑0.8} & 24.8\scriptsize{↓0.3} & 43.9\scriptsize{↓1.6} \\

\textsc{CCU} & Conversational & Coarse-Grain & Unstructured 
& \underline{51.0}\scriptsize{↑1.2} & 26.6\scriptsize{↓0.2} & 69.4\scriptsize{↓1.6} & \underline{\textbf{25.8}}\scriptsize{↑0.1} & 43.2\scriptsize{↓0.1} \\

\midrule
\textsc{Wiki} & Authoritative & Flexible & Unstructured 
& 49.2 & 31.5 & 83.0 & 25.8 & 47.4 \\

\textsc{GenRead mix} & Authoritative & Flexible & Unstructured 
& \textbf{51.4} & 31.9 & \textbf{84.2} & 25.6 & \textbf{48.3} \\
\bottomrule
\end{tabular}
}
\caption{Llama3.1-8b performance comparison across different generated document types combined with Wiki documents using mix with style transformation strategy for various tasks. Superscript values indicate the difference from Table~\ref{tab:rq2_full_llama3.1-8b}, with increases marked by an upward arrow (↑) and decreases by a downward arrow (↓). The best overall performance for each dataset is highlighted in \textbf{bold}, and the best performance within the category of Self-Docs with the same mix strategy is \underline{underlined}.}
\label{tab:rq3_full_direct_mix_with_style_transformed_wiki_llama3.1-8b}
\end{table*}

\begin{table*}[ht]
\vspace{-1mm}
\centering
\resizebox{1\textwidth}{!}{
    \begin{tabular}{c|c|cccc|c}
        \toprule
        \textbf{Meta Function} & \textbf{Dimension} 
        & \textbf{\textsc{TQA}} & \textbf{\textsc{HotpotQA}} & \textbf{\textsc{FEVER}} & \textbf{\textsc{ELI5}} & \textbf{Average} \\
        \midrule
        \multirow{2}{*}[0em]{\textsc{Ideational}} 
                           & \textsc{Fine-Grain}      
                           & 47.1\scriptsize{↓2.2} & \underline{30.4}\scriptsize{↓1.1} & 72.8\scriptsize{↓0.3} & 23.3\scriptsize{↑0.4} & 43.4\scriptsize{↓0.8} \\
                           & \textsc{Coarse-Grain}    
                           & \underline{48.4}\scriptsize{↓1.6} & 28.2\scriptsize{↓1.9} & \underline{73.7}\scriptsize{↑1.5} & \underline{24.0}\scriptsize{↓0.1} & \underline{43.6}\scriptsize{↓0.5} \\
        \midrule
        \multirow{2}{*}[0em]{\textsc{Interpersonal}} 
                           & \textsc{Authoritative}   
                           & 46.7\scriptsize{↓2.2} & 29.3\scriptsize{↓1.9} & \underline{75.5}\scriptsize{↑0.1} & 22.4\scriptsize{↓0.6} & 43.5\scriptsize{↓1.1} \\
                           & \textsc{Conversational}  
                           & \underline{48.8}\scriptsize{↓1.6} & 29.3\scriptsize{↓1.2} & 71.0\scriptsize{↓0.8} & \underline{24.9}\scriptsize{↑0.2} & 43.5\scriptsize{↓0.9} \\
        \midrule
        \multirow{2}{*}[0em]{\textsc{Textual}} 
                           & \textsc{Structured}      
                           & 46.8\scriptsize{↓2.5} & \underline{29.9}\scriptsize{↓2.9} & \underline{74.4}\scriptsize{↑1.1} & 23.5\scriptsize{↑0.4} & \underline{43.7}\scriptsize{↓0.9} \\
                           & \textsc{Unstructured}    
                           & \underline{48.6}\scriptsize{↓1.2} & 28.7\scriptsize{↓0.6} & 72.1\scriptsize{↓2.1} & \underline{23.9}\scriptsize{↓0.4} & 43.3\scriptsize{↓1.1} \\
        \bottomrule
    \end{tabular}
}
\caption{Llama3.1-8b aggregated performance results across different dimensions of the metafunctions (Ideational, Interpersonal, and Textual) on various datasets. Superscript values indicate the difference from Table~\ref{tab:rq2_agg_llama3.1-8b}, with increases marked by an upward arrow (↑) and decreases by a downward arrow (↓). The best performance within each metafunction dimension is \underline{underlined}.}
\vspace{-1mm}
\label{tab:rq3_agg_direct_mix_with_style_transformed_wiki_llama3.1-8b}
\end{table*}


\begin{table*}[h!]
\centering
\small
\resizebox{\textwidth}{!}{
\begin{tabular}{c|c|c|c|c|c|c|c|c}
\toprule
\textbf{Docs Type} & \textbf{Interpersonal} & \textbf{Ideational} & \textbf{Textual} & \textbf{\textsc{TQA}} & \textbf{\textsc{HotpotQA}} & \textbf{\textsc{FEVER}} & \textbf{\textsc{ELI5}} & \textbf{Average} \\
\midrule
\textsc{AFS} & Authoritative & Fine-Grain & Structured  
& 55.8 & 40.0 & 88.6 & 22.3 & 51.7 \\

\textsc{AFU} & Authoritative & Fine-Grain & Unstructured 
& 57.8 & 40.5 & \underline{\textbf{91.8}} & 22.4 & 53.1 \\

\textsc{ACS} & Authoritative & Coarse-Grain & Structured 
& 58.0 & 40.6 & 89.0 & 22.7 & 52.6 \\

\textsc{ACU} & Authoritative & Coarse-Grain & Unstructured 
& 58.8 & \underline{\textbf{41.4}} & 91.2 & 22.9 & 53.6 \\

\textsc{CFS} & Conversational & Fine-Grain & Structured 
& 48.8 & 32.0 & 72.6 & 23.1 & 44.1 \\

\textsc{CFU} & Conversational & Fine-Grain & Unstructured 
& 51.4 & 30.3 & 70.8 & 24.7 & 44.3 \\

\textsc{CCS} & Conversational & Coarse-Grain & Structured 
& 59.0 & 40.5 & 88.8 & 25.1 & 53.4 \\

\textsc{CCU} & Conversational & Coarse-Grain & Unstructured 
& \underline{\textbf{60.4}} & 40.6 & 89.8 & \underline{\textbf{25.6}} & 54.1 \\

\midrule
\textsc{Wiki} & Authoritative & Flexible & Unstructured 
& 52.8 & 38.3 & 91.0 & \textbf{25.6} & 51.9 \\

\bottomrule
\end{tabular}
}
\caption{GPT4o-Mini performance comparison of various document types across tasks. Each document type is categorized based on its interpersonal, ideational, and textual functions. 
The best overall performance for each dataset is highlighted in \textbf{bold}, and the best performance within the category of Self-Docs is \underline{underlined}.}
\label{tab:rq2_full_gpt4o-mini}
\end{table*}

\begin{table*}[h]
\vspace{-1mm}
\centering
\resizebox{1\textwidth}{!}{
    \begin{tabular}{c|c|cccc|c}
        \toprule
        \textbf{Meta Function} & \textbf{Dimension} & \textbf{\textsc{TQA}} & \textbf{\textsc{HotpotQA}} & \textbf{\textsc{FEVER}} & \textbf{\textsc{ELI5}} & \textbf{Average} \\
        \midrule
        \multirow{2}{*}[0em]{\textsc{Ideational}} 
                           & \textsc{Fine-Grain}      
                           & 53.5 & 35.7 & 81.0 & 23.1 & 48.3 \\
                           & \textsc{Coarse-Grain}    
                           & \underline{59.1} & \underline{40.8} & \underline{89.7} & \underline{24.1} & \underline{53.4} \\
        \midrule
        \multirow{2}{*}[0em]{\textsc{Interpersonal}} 
                           & \textsc{Authoritative}   
                           & \underline{57.6} & \underline{40.6} & \underline{90.2} & 22.6 & \underline{52.7} \\
                           & \textsc{Conversational}  
                           & 54.9 & 35.9 & 80.5 & \underline{24.6} & 49.0 \\
        \midrule
        \multirow{2}{*}[0em]{\textsc{Textual}} 
                           & \textsc{Structured}      
                           & 55.4 & \underline{38.3} & 84.8 & 23.3 & 50.4 \\
                           & \textsc{Unstructured}    
                           & \underline{57.1} & 38.2 & \underline{85.9} & \underline{23.9} & \underline{51.3} \\
        \bottomrule
    \end{tabular}
}
\caption{GPT4o-Mini performance comparison of various document types across tasks. Each document type is categorized based on its interpersonal, ideational, and textual functions. 
The best overall performance for each dataset is highlighted in \textbf{bold}, and the best performance within the category of Self-Docs is \underline{underlined}.}
\label{tab:rq2_agg_gpt4o-mini}
\end{table*}

\begin{table*}[h]
\centering
\small
\resizebox{\textwidth}{!}{
\begin{tabular}{c|c|c|c|c|c|c|c|c}
\toprule
\textbf{Type} & \textbf{Interpersonal} & \textbf{Ideational} & \textbf{Textual} 
& \textbf{\textsc{TQA}} & \textbf{\textsc{HotpotQA}} & \textbf{\textsc{FEVER}} & \textbf{\textsc{ELI5}} & \textbf{Average} \\
\midrule
\textsc{AFS} & Authoritative & Fine-Grain & Structured  
& 57.0\scriptsize{↑1.2} & 41.0\scriptsize{↑1.0} & 91.0\scriptsize{↑2.4} & 22.6\scriptsize{↑0.3} & 52.9\scriptsize{↑1.2} \\

\textsc{AFU} & Authoritative & Fine-Grain & Unstructured 
& 58.6\scriptsize{↑0.8} & 40.9\scriptsize{↑0.4} & 91.0\scriptsize{↓0.8} & 22.6\scriptsize{↑0.2} & 53.3\scriptsize{↑0.2} \\

\textsc{ACS} & Authoritative & Coarse-Grain & Structured 
& 59.8\scriptsize{↑1.8} & 39.7\scriptsize{↓0.9} & 91.2\scriptsize{↑2.2} & 23.1\scriptsize{↑0.4} & 53.5\scriptsize{↑0.9} \\

\textsc{ACU} & Authoritative & Coarse-Grain & Unstructured 
& 57.2\scriptsize{↓1.6} & 41.9\scriptsize{↑0.5} & 91.4\scriptsize{↑0.2} & 23.2\scriptsize{↑0.3} & 53.4\scriptsize{↓0.2} \\

\textsc{CFS} & Conversational & Fine-Grain & Structured 
& 58.4\scriptsize{↑9.6} & 40.1\scriptsize{↑8.1} & \underline{\textbf{91.8}}\scriptsize{↑19.2} & 24.5\scriptsize{↑1.4} & 53.7\scriptsize{↑9.6} \\

\textsc{CFU} & Conversational & Fine-Grain & Unstructured 
& 58.6\scriptsize{↑7.2} & 41.7\scriptsize{↑11.4} & 90.2\scriptsize{↑19.4} & 25.1\scriptsize{↑0.4} & 53.9\scriptsize{↑9.6} \\

\textsc{CCS} & Conversational & Coarse-Grain & Structured 
& 58.8\scriptsize{↓0.2} & 40.5\scriptsize{=0.0} & 90.6\scriptsize{↑1.8} & 25.2\scriptsize{↑0.1} & 53.8\scriptsize{↑0.4} \\

\textsc{CCU} & Conversational & Coarse-Grain & Unstructured 
& \underline{\textbf{61.0}}\scriptsize{↑0.6} & \underline{42.0}\scriptsize{↑1.4} & 91.2\scriptsize{↑1.4} & \underline{\textbf{25.7}}\scriptsize{↑0.1} & \underline{\textbf{55.0}}\scriptsize{↑0.9} \\

\midrule
\textsc{Wiki} & Authoritative & Flexible & Unstructured 
& 52.8 & 38.3 & 91.0 & 25.6 & 51.9 \\

\textsc{GenRead mix} & Authoritative & Flexible & Unstructured 
& 57.0 & \textbf{43.1} & 90.0 & 24.2 & 53.6 \\

\bottomrule
\end{tabular}
}
\caption{GPT4o-Mini performance comparison across different generated document types combined with Wiki documents for various tasks. Superscript values indicate the difference from Table~\ref{tab:rq2_full_gpt4o-mini}, with increases marked by an upward arrow (↑) and decreases by a downward arrow (↓). The best overall performance for each dataset is highlighted in \textbf{bold}, and the best performance within the category of Self-Docs with the same mix strategy is \underline{underlined}.}
\label{tab:rq3_full_direct_mix_with_raw_wiki_gpt4o-mini}
\end{table*}

\begin{table*}[ht]
\vspace{-1mm}
\centering
\resizebox{1\textwidth}{!}{
    \begin{tabular}{c|c|cccc|c}
        \toprule
        \textbf{Meta Function} & \textbf{Dimension} 
        & \textbf{\textsc{TQA}} & \textbf{\textsc{HotpotQA}} & \textbf{\textsc{FEVER}} & \textbf{\textsc{ELI5}} & \textbf{Average} \\
        \midrule
        \multirow{2}{*}[0em]{\textsc{Ideational}} 
                           & \textsc{Fine-Grain}      
                           & 58.2\scriptsize{↑4.7} & 40.9\scriptsize{↑5.2} & 91.0\scriptsize{↑10.0} & 23.7\scriptsize{↑0.6} & 53.5\scriptsize{↑5.2} \\
                           & \underline{\textsc{Coarse-Grain}}    
                           & 59.2\scriptsize{↑0.1} & 41.0\scriptsize{↑0.2} & 91.1\scriptsize{↑1.4} & 24.3\scriptsize{↑0.2} & 53.9\scriptsize{↑0.5} \\
        \midrule
        \multirow{2}{*}[0em]{\textsc{Interpersonal}} 
                           & \textsc{Authoritative}   
                           & 58.2\scriptsize{↑0.6} & 40.9\scriptsize{↑0.3} & 91.2\scriptsize{↑1.0} & 22.9\scriptsize{↑0.3} & 53.3\scriptsize{↑0.6} \\
                           & \underline{\textsc{Conversational}}  
                           & 59.2\scriptsize{↑4.3} & 41.1\scriptsize{↑5.2} & 91.0\scriptsize{↑10.5} & 25.1\scriptsize{↑0.5} & 54.1\scriptsize{↑5.1} \\
        \midrule
        \multirow{2}{*}[0em]{\textsc{Textual}} 
                           & \textsc{Structured}      
                           & 58.5\scriptsize{↑3.1} & 40.3\scriptsize{↑2.0} & 91.2\scriptsize{↑6.4} & 23.9\scriptsize{↑0.6} & 53.5\scriptsize{↑3.1} \\
                           & \underline{\textsc{Unstructured}}    
                           & 58.9\scriptsize{↑1.8} & 41.6\scriptsize{↑3.4} & 91.0\scriptsize{↑5.1} & 24.2\scriptsize{↑0.3} & 53.9\scriptsize{↑2.6} \\
        \bottomrule
    \end{tabular}
}
\caption{GPT4o-Mini aggregated performance results across different dimensions of the metafunctions (Ideational, Interpersonal, and Textual) on various datasets. Superscript values indicate the difference from Table~\ref{tab:rq2_agg_gpt4o-mini}, with increases marked by an upward arrow (↑) and decreases by a downward arrow (↓). The best performance within each metafunction dimension is \underline{underlined}.}
\vspace{-1mm}
\label{tab:rq3_agg_direct_mix_with_raw_wiki_gpt4o-mini}
\end{table*}

\begin{table*}[h]
\centering
\resizebox{\textwidth}{!}{
\begin{tabular}{c|c|c|c|c|c|c|c|c}
\toprule
\textbf{Docs Type} & \textbf{Interpersonal} & \textbf{Ideational} & \textbf{Textual}
& \textbf{\textsc{TQA}} & \textbf{\textsc{HotpotQA}} & \textbf{\textsc{FEVER}} & \textbf{\textsc{ELI5}} & \textbf{Average} \\
\midrule
\textsc{AFS} & Authoritative & Fine-Grain & Structured
& 54.4\scriptsize{↓1.4} & 41.0\scriptsize{↑1.0} & 89.8\scriptsize{↑1.2} & 22.4\scriptsize{↑0.1} & 51.9\scriptsize{↑0.2} \\

\textsc{AFU} & Authoritative & Fine-Grain & Unstructured
& 57.6\scriptsize{↓0.2} & 40.3\scriptsize{↓0.2} & \underline{\textbf{92.2}}\scriptsize{↑0.4} & 22.5\scriptsize{↑0.1} & 53.2\scriptsize{↑0.1} \\

\textsc{ACS} & Authoritative & Coarse-Grain & Structured
& 58.8\scriptsize{↑0.8} & 39.9\scriptsize{↓0.7} & 90.8\scriptsize{↑1.8} & 22.9\scriptsize{↑0.2} & 53.1\scriptsize{↑0.5} \\

\textsc{ACU} & Authoritative & Coarse-Grain & Unstructured
& 58.0\scriptsize{↓0.8} & 40.4\scriptsize{↓1.0} & 90.8\scriptsize{↓0.4} & 22.8\scriptsize{↓0.1} & 53.0\scriptsize{↓0.6} \\

\textsc{CFS} & Conversational & Fine-Grain & Structured
& 56.8\scriptsize{↑8.0} & 40.0\scriptsize{↑8.0} & 91.2\scriptsize{↑18.6} & 24.1\scriptsize{↑1.0} & 53.0\scriptsize{↑8.9} \\

\textsc{CFU} & Conversational & Fine-Grain & Unstructured
& 58.8\scriptsize{↑7.4} & \underline{41.5}\scriptsize{↑11.2} & 90.2\scriptsize{↑19.4} & 24.8\scriptsize{↑0.1} & 53.8\scriptsize{↑9.5} \\

\textsc{CCS} & Conversational & Coarse-Grain & Structured
& 60.6\scriptsize{↑1.6} & 40.6\scriptsize{↑0.1} & 90.6\scriptsize{↑1.8} & 25.1\scriptsize{=0.0} & 54.3\scriptsize{↑0.9} \\

\textsc{CCU} & Conversational & Coarse-Grain & Unstructured
& \underline{\textbf{61.0}}\scriptsize{↑0.6} & 41.4\scriptsize{↑0.8} & 91.4\scriptsize{↑1.6} & \underline{25.4}\scriptsize{↓0.2} & \underline{\textbf{54.8}}\scriptsize{↑0.7} \\

\midrule
\textsc{Wiki} & Authoritative & Flexible & Unstructured
& 52.8 & 38.3 & \textbf{91.0} & \textbf{25.6} & 51.9 \\

\textsc{GenRead mix} & Authoritative & Flexible & Unstructured
& 57.0 & \textbf{43.1} & 90.0 & 24.2 & 53.6 \\

\bottomrule
\end{tabular}
}
\caption{GPT4o-Mini performance comparison across different generated document types combined with Wiki documents using mix with style transformation strategy for various tasks. Superscript values indicate the difference from Table~\ref{tab:rq2_full_gpt4o-mini}, with increases marked by an upward arrow (↑) and decreases by a downward arrow (↓). The best overall performance for each dataset is highlighted in \textbf{bold}, and the best performance within the category of Self-Docs is \underline{underlined}.}
\label{tab:rq3_full_direct_mix_with_style_transformed_wiki_gpt4o-mini}
\end{table*}

\begin{table*}[ht]
\vspace{-1mm}
\centering
\resizebox{1\textwidth}{!}{
    \begin{tabular}{c|c|cccc|c}
        \toprule
        \textbf{Meta Function} & \textbf{Dimension} 
        & \textbf{\textsc{TQA}} & \textbf{\textsc{HotpotQA}} & \textbf{\textsc{FEVER}} & \textbf{\textsc{ELI5}} & \textbf{Average} \\
        \midrule
        \multirow{2}{*}[0em]{\textsc{Ideational}} 
                           & \textsc{Fine-Grain}      
                           & 56.9\scriptsize{↑3.4} & \underline{40.7}\scriptsize{↑5.0} & \underline{90.9}\scriptsize{↑9.9} & 23.5\scriptsize{↑0.4} & 53.0\scriptsize{↑4.7} \\
                           & \underline{\textsc{Coarse-Grain}}    
                           & \underline{59.6}\scriptsize{↑0.5} & 40.6\scriptsize{↓0.2} & \underline{90.9}\scriptsize{↑1.2} & \underline{24.1}\scriptsize{=0.0} & \underline{53.8}\scriptsize{↑0.4} \\
        \midrule
        \multirow{2}{*}[0em]{\textsc{Interpersonal}} 
                           & \textsc{Authoritative}   
                           & 57.2\scriptsize{↓0.4} & 40.4\scriptsize{↓0.2} & \underline{90.9}\scriptsize{↑0.7} & 22.7\scriptsize{↑0.1} & 52.8\scriptsize{↑0.1} \\
                           & \underline{\textsc{Conversational}}  
                           & \underline{59.3}\scriptsize{↑4.4} & \underline{40.9}\scriptsize{↑5.0} & \underline{90.9}\scriptsize{↑10.4} & \underline{24.9}\scriptsize{↑0.3} & \underline{54.0}\scriptsize{↑5.0} \\
        \midrule
        \multirow{2}{*}[0em]{\textsc{Textual}} 
                           & \textsc{Structured}      
                           & 57.7\scriptsize{↑2.3} & 40.4\scriptsize{↑2.1} & 90.6\scriptsize{↑5.8} & 23.6\scriptsize{↑0.3} & 53.1\scriptsize{↑2.7} \\
                           & \underline{\textsc{Unstructured}}    
                           & \underline{58.9}\scriptsize{↑1.8} & \underline{40.9}\scriptsize{↑2.7} & \underline{91.2}\scriptsize{↑5.3} & \underline{23.9}\scriptsize{=0.0} & \underline{53.7}\scriptsize{↑2.4} \\
        \bottomrule
    \end{tabular}
}
\caption{GPT4o-Mini aggregated performance results across different dimensions of the metafunctions (Ideational, Interpersonal, and Textual) on various datasets. Superscript values indicate the difference from Table~\ref{tab:rq2_agg_gpt4o-mini}, with increases marked by an upward arrow (↑) and decreases by a downward arrow (↓). The best performance within each metafunction dimension is \underline{underlined}.}
\vspace{-1mm}
\label{tab:rq3_agg_direct_mix_with_style_transformed_wiki_gpt4o-mini}
\end{table*}

\end{document}